\documentclass[10pt,final,journal]{IEEEtran}
\usepackage{fancyhdr}
\usepackage[switch]{lineno} 
\usepackage{tabularx} 

\usepackage{stmaryrd}

\usepackage{cite}

\ifCLASSINFOpdf
\usepackage[pdftex]{graphicx}
\else
\fi
\usepackage{amsfonts}

\usepackage{amsmath}
\usepackage{algorithm}
\usepackage{algpseudocode}
\ifCLASSOPTIONcompsoc
 \usepackage[caption=false,font=normalsize,labelfont=sf,textfont=sf]{subfig}
\else
 \usepackage[caption=false,font=footnotesize]{subfig}
\fi
\begin{document}
%
\title{BLAS: Broadcast Relative Localization and \\ Clock Synchronization 
for Dynamic Dense \\ Multi-Agent Systems}



\author{Qin~Shi,
        Xiaowei~Cui,
        Sihao~Zhao,
        Shuang~Xu,
        and~Mingquan~Lu
\thanks{Q. Shi, X. Cui, S. Zhao and S. Xu are with the Department of Electronic Engineering,
Tsinghua University, Beijing 100084, P. R. China (sqn175@gmail.com; cxw2005@tsinghua.edu.cn;
zsh{\_}thu@mail.tsinghua.edu.cn; xushuang1109@163.com).}
\thanks{M. Lu is with is with Department of Electronic Engineering, Tsinghua University
and Beijing National Research Center for Information Science and Technology, Beijing, 100084, P. R. China
(lumq@tsinghua.edu.cn).}}

%




\maketitle

\begin{abstract}
The spatiotemporal information plays crucial roles in a multi-agent system (MAS). 
However, for a highly dynamic and dense MAS in unknown environments,
estimating its spatiotemporal states is a difficult problem.
In this paper, we present BLAS:  
a wireless broadcast relative localization and clock synchronization system to 
address these challenges.
Our BLAS system exploits a broadcast architecture, under which 
a MAS is categorized into parent agents that 
broadcast wireless packets and child agents that are passive receivers, to 
reduce the number of required 
packets among agents for relative localization and clock synchronization.
We first propose an asynchronous broadcasting and passively receiving (ABPR) protocol.
The protocol schedules the broadcast of parent agents using a 
distributed time division multiple access (D-TDMA) scheme
and delivers inter-agent information used for joint 
relative localization and clock synchronization. 
We then present distributed state estimation approaches in parent and child agents that
utilize the broadcast inter-agent information for joint estimation of spatiotemporal states. 
The simulations and real-world experiments based on ultra-wideband (UWB)
illustrate that our proposed BLAS cannot only
enable accurate, high-frequency and real-time estimation of relative position and clock parameters
but also support theoretically an unlimited number of agents.
\end{abstract} 



%
\IEEEpeerreviewmaketitle

\section{Introduction}
%
%
%
%
\IEEEPARstart{A}{long} with the artificial intelligence tendency and the rapid 
development of electronic, sensor and control technologies, agents such as 
unmanned aerial vehicles (UAV) and unmanned ground vehicles (UGV) have been made 
possible to autonomously move in harsh environments and operate remotely without 
human intervention. Because of their flexibility, reconfigurability, and intelligence, 
there has been a tremendous increase in the use of such intelligent agents in tasks
like target search and track \cite{pitre2012uav}, 
photovoltaic plant inspection \cite{addabbo2018uav}, 
and planet exploration \cite{maimone2007two}. 
Over the past few years, systems containing only a single agent have been intensively investigated 
and used. However, the capability of a single agent is limited.
In the future applications with growing complexity and uncertainty, deploying a team 
of small agents, namely a multi-agent system (MAS) is anticipated to offer better capabilities 
beyond only 
a single agent, in terms of robustness, scalability, and effectiveness.

Realizing this vision will require MASs to overcome some new unique 
challenges. Among these, real-time precise relative localization and 
clock synchronization are two key challenges.
Relative localization is the process of determining the multi-agent dynamic topology, and 
clock synchronization provides a common time reference for distributed agents. 
The spatiotemporal determination is crucial for MASs to perform 
basic operations: 
\textit{1) Coordination}: Spatiotemporal coordination between agents is necessary 
for a MAS to collaboratively carry out tasks effectively.
The difficulty of the coordination operations strongly 
lies in the knowledge of the relative position and clock parameters among the agents. 
If these parameters are known, effective coordination can be achieved by involving 
the state of all the agents at the same time \cite{dai2013optimal}.
\textit{2) Data fusion}: Data fusion is an essential operation to share and integrate 
sensory data collected by distributed agents. To fuse in a meaningful way, these data are 
coupled with time and location information. However, the information is expressed with 
respect to each agent's local reference and needs to be transformed to a common reference. 
Such a transformation requires relative localization and clock synchronization of MASs.
Moreover, the inter-agent collision avoidance, collaborative environment mapping, and
formation control etc. also require spatiotemporal information.
Therefore, relative localization and clock synchronization have become pressing issues in the design and 
advancement of MASs. 

Using external spatiotemporal reference is a practical solution. 
External reference options, such as the well-known Global Positioning System (GPS), 
optical motion capture systems, and the trendy ultra-wideband (UWB) anchors, have powered  
MASs for recreational or professional use
\cite{vasarhelyi2018optimized, kushleyev2013towards, hamer2018self}.
Terrestrial wireless communications and the broadcast signals of opportunity (SOPs), such as 
cellular signals \cite{muller2016statistical, khalife2018navigation, shamaei2018exploiting, yang2014mobile} 
and the signals from Iridium system \cite{pesyna2012constructing}, 
have also been demonstrated to navigate MASs in recent studies. 
In all cases, by comparing absolute position and time relative to the external reference, 
the relative localization and clock synchronization among each agent can be straightforwardly determined. 
However, such references require pre-installed infrastructures, which restricts the 
maneuverability of agents and are not always available 
or reliable in unknown environments. 

To enable future multi-agent applications in any environment, relative localization and clock synchronization must 
rely on inter-agent measurements other than external references \cite{yang2014covariance}.
We notice that in the wireless sensor network (WSN) community, relative localization and clock synchronization are also 
necessary for spatiotemporal information acquisition in each sensor 
node for meaningful sensory data processing.
Typically, the localization of sensor nodes requires inter-node ranging measurements.
Regarding the ranging measurements, a popular choice is to use the time-of-arrival (TOA) metric, 
which relies on precise clock synchronization and time management.
The clock synchronization is typically conducted in a two-way timing message exchange mechanism, which requires 
two-way TOA measurements between nodes \cite{wu2011clock}. 
Accordingly, the 
synergy between relative localization and clock synchronization is obvious. Various methods for 
joint localization and synchronization (JLAS)
in WSN using TOA measurements have been proposed
\cite{denis2006joint, vaghefi2015cooperative, etzlinger2017cooperative}.  

Indeed, a MAS can be treated as a mobile WSN, which motivates us to 
jointly perform relative localization and clock synchronization in MASs using the aforementioned methods in WSNs.
However, the aforementioned methods are only suitable for static (or partially static) 
and sparse WSNs with low updating frequency requirements, which are difficult to 
apply to MASs. For MASs without external spatiotemporal reference support,
we identify the additional three main challenges to the relative localization and clock synchronization problems:

\begin{enumerate}
    \item \textit{High-density}. MASs typically consist of dense small agents 
and act like swarms. Deriving two-way TOA measurements in a round-robin way 
for all agents is time and power consuming. Special considerations on communication protocol
should be given in developing relative localization and clock synchronization methods, thus supporting theoretically
unlimited agents simultaneously.
    \item \textit{High-maneuverability}. MASs are characterized by its high mobility and dynamics,
    e.g., with a maximum speed of 50 m/s.
The collaboration control operations need high-frequency and real-time (e.g., 100 Hz) 
input of 
spatiotemporal information of all agents. However, conventional relative localization and clock synchronization methods 
in WSNs typically only provide 
one-shot estimates during a long period.
    \item \textit{SWaP constraints}. Last but not least, small agents usually come with size, 
weight, and power (SWaP) constraints, limiting their sensory payload and onboard resources 
for computation, signal processing, and communication. 
\end{enumerate} 

\begin{figure}
    \centering
    \includegraphics[width=0.9\linewidth]{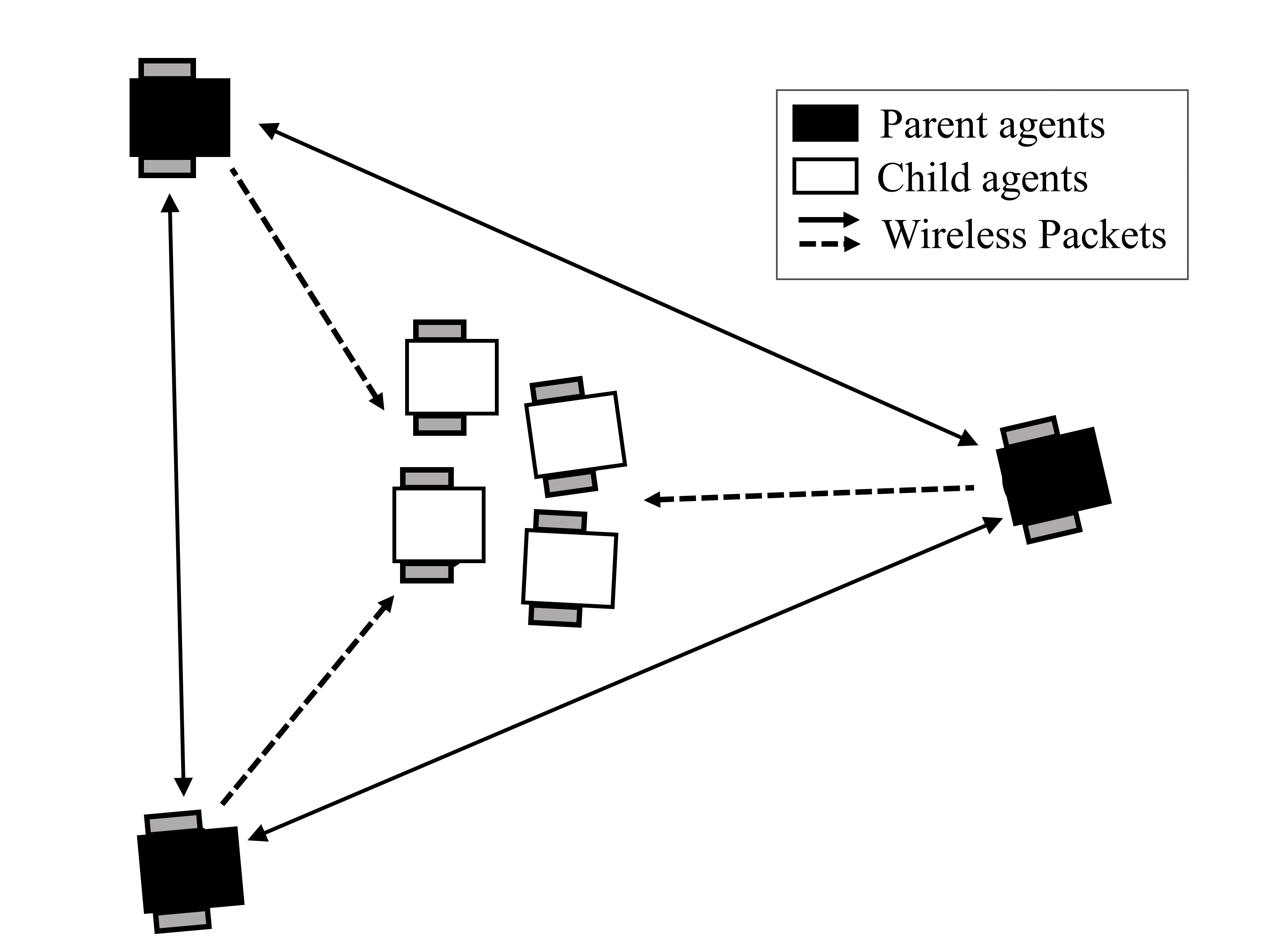} 
    \caption{A MAS is categorized into parent and child agents. Parent agents act like moving 
    beacons by periodically broadcasting wireless radio packets, and child agents only passively receive the packets.
    }
    \label{fig:BLAS}
\end{figure}
 
To address all these challenges, we present BLAS, a precise, effective and reliable relative localization 
and clock synchronization system based on wireless broadcast architecture. 
The broadcast architecture is introduced to reduce the inter-agent communication overhead devoted to 
localization and synchronization in the networked MAS. Under the architecture, a MAS is 
categorized as parent and child agents as shown in Fig. \ref{fig:BLAS}.
Parent agents act like moving beacons by periodically broadcasting wireless 
packets and child agents only passively receive the on-air packets.
In BLAS, we first propose an asynchronous broadcasting and passively receiving (ABPR) communication protocol. 
The protocol serves two purposes: 
it schedules the broadcast of the parent agents using a distributed time division multiple access (D-TDMA) scheme
and it also specifies the information embedded in broadcast packets, which is dedicated to 
joint relative localization and clock synchronization for both parent and child agents.
Using the protocol, 
a two-way link can then be established between any pair of parent agents. 
We then present distributed state estimation approaches in parent agents to establish a 
moving spatiotemporal reference frame using the two-way packets.
A distributed approach in child agents  
is also presented to jointly estimate the spatiotemporal states 
with respect to the pre-defined moving spatiotemporal reference,
by only passively receiving the broadcast packets.

We note that BLAS only requires the broadcast packets to yield the inter-agent TOA measurements, 
thus can be considered as an 
enabling technique for MASs operating in any environment.
The main contributions of this paper are summarized as follows:
\begin{itemize}
    \item A novel wireless broadcast architecture based on which a relative localization and clock synchronization system 
    is developed for dynamic and dense MASs with little resource requirements. 
    \item A versatile communication protocol that simultaneously supports joint relative localization and clock synchronization 
    for both parent and child agents.
    \item Real-time implementation of the system utilizing the off-the-shelf low-cost and 
    light-weight hardware. 
\end{itemize} 

The rest of the paper is organized as follows. We first review the related literature on 
relative localization and clock synchronization in Section II. We then give our model 
formulations in Section III. In Section IV, we present our proposed ABPR protocol. We
then present the joint relative localization and clock synchronization methods of parent 
agents and child agents in Section V and Section VI. 
Finally, simulations and real-world experiments are evaluated in Section VII and Section VIII. 
Section IX concludes the paper with comments on future works.

\section{Related Work}
This section gives a concise review of related literature on relative localization and 
clock synchronization. These problems have been widely studied in various areas, including 
wireless sensor networks, robotics, and signal processing. The related scholarly works are 
extensive and hence we just focus on the most relevant papers.

\subsection{Relative Localization}
Relative localization refers to the problem of perceiving the position of other 
agents in the surroundings. The perception requires onboard sensors to provide inter-agent 
measurements. Numerous implementations have been developed utilizing 
different onboard sensors. According to the sensory modality, we can categorize these 
implementations into two groups: indirect and direct relative localization. 

Indirect relative localization in MASs relies on optical 
sensors and/or laser scanners that can extract the environment-specific 
static features and landmarks. In these agents, the relative position is indirectly 
derived from observations of the same features and landmarks. For example, 
cameras are mounted in three unmanned aerial vehicles (UAVs) to identify the 
common objects in the environment, and the relative position between UAVs is 
calculated from these correspondences \cite{merino2006vision}. An alternative 
method exchanges the laser range scans between pairs of robots and determines the 
relative positions by estimating the overlap of their partial maps \cite{ko2003practical}.
However, observation of the same scene is occasional and with great uncertainty.
In the robotics community, the relative localization is always coupled with the 
environment mapping process, 
which leads to the multi-robot Simultaneous Localization and Mapping (SLAM) 
problem.  
The precise and continuous relative position estimates are derived by localizing 
the robots in a common global incremental map. Recent research has experimentally 
shown that multi-robot SLAM utilizing 
cameras \cite{zou2013coslam} or 
laser scanners \cite{howard2006multi}
can produce accurate relative position estimates.
Such an approach builds on high-resolution cameras or high-quality laser scanners and 
on computationally expensive algorithms, such as smoothing \cite{zou2013coslam} and
particle filtering \cite{howard2006multi}, to reduce uncertainty. 
Consequently, SLAM based relative localization requires heavy sensor payload and a powerful 
processor which is not always available on SWaP constrained agents. 

Alternatively, direct relative localization is pursued using direct inter-agent 
measurements. For example, an audio-based approach directly perceives the neighbors 
by exploiting the direction and intensity of the sound emitted by its neighboring 
agents' onboard piezo \cite{basiri2014audio}. However, the accuracy is relatively low 
due to the sound noise of engines and/or propellers \cite{sinibaldi2013experimental}. 
High-frequency-modulated infrared have been used to enable nearby agents 
to determine the relative position of the sender 
with a fast update rate \cite{pugh2009fast}, however, the relatively short operating range of a
few meters is not suitable for large-scale environments.
A vision-based approach detects the relative position of neighbors 
by recognizing the unique localization image patterns attached to 
neighboring agents \cite{krajnik2014practical}. However, the constraints specified 
by camera shutter frequency and field of view prevents its use for high dynamic 
agents. The natural weakness of visual methods is obvious at 
night or in light-changing environments. 
In all cases, the inter-agent measurements are derived between each pair of agents
which produces redundancy and need more measuring time.
Therefore, scalable implementations for high-density agents are impractical. 

A much preferable approach for scalable relative localization is to communicate the inter-agent
measurements through the networked MASs, to which we refer as mutual relative 
localization \cite{shames2011cooperative}. Hence, there is a need for MASs to establish a communication 
link between the agents as well as to produce inter-agent measurements. 
UWB is a promising sensory technology to endow MASs with 
the ability of simultaneous communication and range measuring \cite{win2000ultra, lee2002ranging}. 
The fine temporal resolution and robustness to multipath enable
reliable and precise direct inter-agent ranging measurements.  
Additionally, the advent of small, low-cost and low-power UWB transceiver chips
makes it a suitable sensory choice for small SWaP constrained agents. 
To this end, our work builds on the UWB technology to realize mutual relative localization
in MASs. To the best of our knowledge, no studies have demonstrated a 
UWB based mutual relative localization for high-density and high-maneuverability agents.
Of particular relevance is the numerous works for the mutual relative localization of 
static agents in the WSN 
community \cite{ahmed2005sharp, patwari2005locating, buehrer2018collaborative}. Closest in spirit to our approach
is the work of "SHARP" \cite{ahmed2005sharp}, which first localized several reference nodes then localized
the rest of the nodes with respect to the reference nodes.
The difference between our work and the works in WSN is that our work 
deals with the dynamic agents with the assumption that all agents are in the 
UWB communication range of each other.

\subsection{Clock Synchronization}
In MASs, there is no global clock. Each agent has its own
internal clock that reads its own time notion. Although they might be 
synchronized when they start, the frequency of the oscillators in each 
agent slightly differs from each other, thus leading to a drift in the 
notion of time. 
Clock synchronization has been widely investigated in wired networks, 
for example, the well-known Network Time Protocol (NTP) has been 
widely used for time synchronization on the Internet \cite{mills1991internet}.
However, they are not suitable for wireless 
networked MASs which have potential motion, SWaP constraints and signal interference. 

Clock synchronization is also a widely studied topic in the WSN 
community \cite{wu2011clock}.
In WSNs, it can be achieved by sending a sequence of timing messages 
to the receiver sensors. The receiver sensors record the receiving time 
of the broadcasting messages and extract the embedded broadcast time 
information for synchronization processing.
Generally, there are two different approaches according to the message
transferring protocol: sender-receiver synchronization (SRS) and 
receiver-receiver synchronization (RRS).
SRS is performed using the traditional two-way message exchange mechanism to 
synchronize the receivers with the sender,
such as Romer's protocol \cite{romer2001time}.
Instead of interacting with the sender, 
RRS compares the receiving time of a same broadcast timing message among a set 
of receiving nodes to synchronize the receivers. A typical approach is
the Reference-Broadcast Synchronization (RBS) \cite{elson2002fine}.
However, to achieve network-wide synchronization, SRS and RRS approaches require
a numerous number of timing messages, thus are time and power consuming, and 
increase communication overheads in dense networks.
The Pairwise Broadcast Synchronization (PBS) is a promising way 
to reduce the overall consumption, which innovatively combines the RRS 
and SRS protocols \cite{noh2008new}. In PBS, groups of nodes are 
synchronized by only passively receiving timing messages from  
parent nodes which perform a pairwise synchronization using SRS mechanism.
This mainly motivates our work in this paper to tackle the 
high-density challenge of MASs. 
However, PBS assumes the distances between some nodes are known, which is not 
suitable for dynamic networks, e.g., networked MASs.

\subsection{JLAS}
The need to recover the distances between nodes for clock synchronization
has driven the development of JLAS algorithms. As we have mentioned in the previous 
section, the distances can be recovered through TOA measurements, which 
require fine clock synchronization between nodes. Consequently, the 
distance based localization problem and the clock synchronization problem
is highly related and can be jointly tackled. An early work \cite{denis2006joint} 
jointly solved the problems by first conducting clock synchronization and then 
localization using the same communication means, which is designed for
static WSNs requiring low updating rate. 
Recently, JLAS have been solved simultaneously
in one step. For example, a joint maximum likelihood
estimator and a least squares (LS) estimator were proposed to determine the 
spatiotemporal information of an unknown sensor node with accurate anchors
available, and a generalized total LS scheme is proposed with inaccurate 
anchors \cite{zheng2010joint}, however, they are computationally expensive. 
A low-complexity solution based on the linearized equations from TOA measurements
and a weighted least square (WLS) criterion have proven to achieve a better 
estimation performance \cite{zhu2010joint}. Approximately efficient and closed-form 
solutions were developed in \cite{wang2015toa}.
However, all these methods require pre-installed static anchors with known positions and 
clock parameters. They have not yet been extended to the case with 
mobile anchors of unknown states. 

An insightful JLAS approach considering unknown anchors' states was proposed in \cite{morales2018information}.
It adopted a collaborative radio SLAM framework to simultaneously estimates the anchors' states 
with the MASs' states. Under the framework, each agent fused its onboard inertial information, 
GPS information, TOA measurements from anchors and inter-agent information from its neighboring agents to 
estimate all the states.
In contrast to this fusion strategy, we choose to separately estimate the states of the mobile anchors
(i.e., parent agents), and the states of each agent (i.e., child agent) using only UWB TOA measurements. 
In this way, the potential algorithm complexity improvement due to inter-agent information 
for a dense MAS can be eliminated.

Our proposed system can be treated like GPS in some aspects. 
JLAS in child agents using TOAs can be considered the same 
process in GPS user end, which was addressed using 
Bancroft algorithm \cite{bancroft1985algebraic}. However, Bancroft assumes the TOAs
are measured concurrently, which is reasonable in GPS due to 
the code division multiple access (CDMA) scheme. For MASs utilizing TDMA scheme, the TOAs 
are measured consecutively. If the minor time difference between the 
consecutive TOAs are not taken into account during joint estimation,
significant error can arise for high-maneuverability agents.

\section{Model Formulation}

\subsection{Problem Statement}
This paper studies the problem of relative localization and clock synchronization 
for a highly dynamic and dense MAS in unknown environments utilizing UWB technology.
We focus on 2D space localization 
due to its practical advantages of easier implementation and experimental evaluation. 
Extension to 3D space of our proposed ideas is not difficult.

Suppose a MAS $\mathcal M$ is composed of $M$ mobile agents and is categorized into $P$ parent agents 
and $C$ child agents. Let 
$\mathcal P=\{1,\cdots,P\}$ denotes the set of parent agents with 
unique identification indices and $\mathcal C=\{P+1,\cdots,P+C\}$ the set of child agents.
The cardinality of the sets, i.e., $P$ and $C$ gives the total number of 
parent agents and child agents in the MAS, and $\mathcal M=\mathcal P \cup \mathcal C$.
Every agent is equipped with an UWB transceiver. 
For parent agents, the UWB transceiver 
is able to broadcast and receive UWB packets. For child agents, it only passively 
receives the on-air packets. We assume that the child agents are in the communication 
range of parent agents and the communication link between parent agents are available all 
the time. This is reasonable considering the maximum UWB communication range is relatively long  
and the topology size of swarmed MASs is relatively small. The goal of this 
paper is to real-time estimate the relative positions 
$\mathbf x_i(t) = [x_i, y_i]^T$ of agent $i \in \mathcal M$ with respect to a reference 
coordinate attached to the reference agent $j$ (usually $j=1$), 
as well as the relative clock parameters $T_i^j(t)$, $\omega_i^j(t)$ with respect to the 
reference agent.

\subsection{Clock Model}
Each UWB transceiver hosts an internal hardware clock, which 
is used to 1) timestamp the receiving and 
broadcasting event of UWB packets; 2) schedule the signal broadcasting; 
3) synchronize the agents' clock. Here we note that the clock synchronization
between the agent and its UWB transceiver is addressed by a wired connection.
For simplification, we refer the clock of an agent as the clock of the corresponding UWB transceiver.
The clock value $T_i(t)$ of an agent $i \in \mathcal M$ 
is then read by counting the oscillations in the hardware clock oscillator which runs at 
a particular frequency, where $t$ is the absolute reference time. For a perfect clock, $T_i(t)=t$.
However, the physical oscillators in each agent slightly differ from others 
due to environment changes, such as ambient temperatures and magnetic field, 
thus leading to frequency instabilities and the variation of clock value \cite{barnes1971characterization}.
Using the terminology in \cite{moon1999estimation}, we denote the clock offset of agent $i$ with 
respect to the perfect clock as $T_i^0(t)$, and the clock skew (the difference in the frequencies) 
of agent $i$ with respect to the perfect clock as $\omega_i^0(t)$.
The clock value of agent $i$ is then given as
\begin{equation} \label{eq:clock model}
        T_i(t) = t + T_i^0(t).
\end{equation}
The agent clock parameters are given by \mbox{$\mathbf c_i^0 = [T_i^0, \omega_i^0]^T$}, 
which consist of its clock offset and clock skew. The agent clock's dynamics can be modeled as \cite{kassas2015receding}
\begin{equation} \label{eq:ideal clk model}
\dot {\mathbf c_i^0(t)} = \mathbf F \mathbf c_i^0(t) + \mathbf n_i(t),
\end{equation}
$$
    \mathbf F = 
    \begin{bmatrix}
        0 & 1 \\ 0 & 0
    \end{bmatrix}, \quad
    \mathbf n_i = 
    \begin{bmatrix}
        n_{T_i^0}\\ n_{\omega_i^0}
    \end{bmatrix},
$$
where $n_{T_i^0}$ and $n_{\omega_i^0}$ are modeled as mutually independent zero-mean white noises.
It is assumed that the collections of $\{n_{T_i^0}\mid i \in \mathcal M\}$ and 
$\{n_{\omega_i^0}\mid i \in \mathcal M\}$ are both 
independent and identically distributed with power spectra $S_{n_T}$ and $S_{n_\omega}$, respectively.

However, there is no global absolute time reference in MASs,
the clock offset and the clock skew of an agent cannot be measured.
Typically, any agent's clock can be chosen as a time reference,
hence we only concern about the relative clock offset and relative clock skew. 
Therefore, we define the relative clock offset and clock skew of agent $i$ with respect to agent $j$
at time $t$, $T_i^j(t)$ and $\omega_i^j(t)$, as follows:
\begin{align} 
    T_i^j(t) &= T_i^0(t) - T_j^0(t),  \label{eq:clock offset}\\
    \omega_i^j(t) &= \omega_i^0(t) - \omega_j^0(t) \label{eq:clock skew}.
\end{align}
If agent $i$ and agent $j$ are perfectly synchronized, then we have:
$T_i^j(t)=0$ and $\omega_i^j(t)=0$. 
It can be readily seen that the relative clock parameters evolve according to:
\begin{equation} \label{eq:relative clk model}
    \dot {\mathbf c_i^j}(t) = \mathbf F \mathbf c_i^j(t) + \mathbf n(t),
\end{equation}
$$
    \mathbf c_i^j = 
    \begin{bmatrix}
        T_i^j \\ \omega_i^j
    \end{bmatrix}, \quad
    \mathbf n = 
    \begin{bmatrix}
        n_{T}\\ n_{\omega}
    \end{bmatrix},
$$
where $n_{T}$ and $n_{\omega}$ are with power spectra $2S_{n_T}$ and $2S_{n_\omega}$, respectively.

\subsection{Measurement Model}
Consider an UWB packet broadcast from a parent agent $j\in \mathcal P$ to an 
arbitrary agent $i \in \mathcal M$, 
the broadcasting time $t_{tx}^{j\shortrightarrow i}$ and
reception time $t_{rx}^{j\shortrightarrow i}$,
both expressed in absolute time reference, are related by:
\begin{equation} \label{eq:raw TOA}
    t_{rx}^{j\shortrightarrow i} = t_{tx}^{j\shortrightarrow i} + \tau_i^j(t_{tx}),
\end{equation}
where $\tau_i^j(t_{tx})$ is the signal delay that can be characterized into several distinct parts:
\begin{equation} \label{eq:sig delay}
 \tau_i^j(t_{tx}) = \tau_j + \tau^{ij}(t_{tx}) + \tau_i,
\end{equation}
where $\tau_j$ is the broadcasting antenna delay of agent $j$, $\tau_i$ 
the receiving antenna delay of agent $i$, and $\tau^{ij}(t_{tx})$ the propagation delay.
The antenna delays are constant and dependent on the antenna design, thus are considered 
as a bias in the TOA measurement \cite{wang2012cramer}.
In this paper, we assume that they are deterministic and can be pre-calibrated.
The propagation delay $\tau^{ij}(t)$ is given by:
$$
    \tau^{ij}(t) = \frac {d^{ij}(t)}{v_c},
$$
where $v_c$ is the light speed, and 
$d^{ij}(t)$ is the distance between agent $i$ and $j$,
$d^{ij}(t)=\Vert \mathbf x_j(t) - \mathbf x_i(t) \Vert$.

Typically, the reception time and broadcasting time are 
measured using individual agents' internal clocks according \mbox{to (\ref{eq:clock model})}.
We denote the measured broadcasting timestamp at 
agent $j$ and the reception timestamp 
at agent $i$ as $T_j(t_{tx}^{j\shortrightarrow i})$ and
$T_i(t_{rx}^{j\shortrightarrow i})$ respectively. 
The TOA measurement is then given by:
\begin{equation} \label{eq:TOA meas}
    \tau(t_{tx}^{j\shortrightarrow i}) = T_i(t_{rx}^{j\shortrightarrow i})
    - T_j(t_{tx}^{j\shortrightarrow i}) + n^{ij}(t_{rx}^{j\shortrightarrow i}),
\end{equation}
where $n^{ij}(t_{rx}^{j\shortrightarrow i})$ denotes the random delay caused by 
estimation errors on the stamping process, such as due to noise and multipath. 
The random delay is modeled as independent and identically distributed 
Gaussian noise \cite{elson2002fine}, $n^{ij}\sim \mathcal N(0,\xi^2)$. 
Since the clocks between agents are not synchronized, using  
(\ref{eq:clock model}), (\ref{eq:clock offset}), (\ref{eq:ideal clk model}) and (\ref{eq:raw TOA})
the TOA measurement has the following form:
$$
    \tau(t_{tx}^{j\shortrightarrow i}) = T_i^j(t_{tx}^{j\shortrightarrow i}) + 
        \tau_i^j(t_{tx}^{j\shortrightarrow i}) + 
        \int_{t_{tx}^{j\shortrightarrow i}}^{t_{rx}^{j\shortrightarrow i}} \dot T_i^0(\tau)d\tau
        + n^{ij}(t_{rx}^{j\shortrightarrow i}).
$$
The first item of the equation corresponds to the relative clock offset
between the broadcasting and the receiving agent at broadcasting time.
The sum of following two items corresponds to the signal 
delay which is expressed in the agent $i$'s clock. 
Practically, the magnitude of the clock skew $\omega_i^0$ is unknown
and can be up to $\pm20$ parts per million (ppm) as 
the IEEE 802.15.4a standard indicates \cite{neirynck2016alternative}.
Consider the worst case $\omega_i^0$ of a constant 20 ppm over the propagation interval, 
the integration part is then approximately $2 \times 10^{-5}\cdot \tau_i^j(t_{tx}^{j\shortrightarrow i})$, 
which is obviously much smaller than $\tau_i^j(t_{tx}^{j\shortrightarrow i})$.
Consequently, by ignoring the integration term, we come to the following approximation of
TOA measurement:
\begin{equation} \label{eq:TOA model}
    \tau(t_{tx}^{j\shortrightarrow i}) \cong T_i^j(t_{tx}^{j\shortrightarrow i})
        + \tau_i^j(t_{tx}^{j\shortrightarrow i}) 
        + n^{ij}(t_{rx}^{j\shortrightarrow i}).
\end{equation}

At this point, we have shown that the TOA measurement
is a function of the clock parameters 
and the physical relative positions of agents
(correspond to the signal delay). Consequently, 
TOA measurements can be used for joint relative localization and clock synchronization.

\section{The ABPR Protocol}
In our proposed BLAS system, 
the broadcasting of UWB packets is only performed by parent agents. 
All the remaining $M-1$ agents 
(including other parent agents and child agents) 
will receive the broadcast packets to derive the TOA measurements for state estimation.
In order to support such an architecture, we employ an ABPR communication protocol, which schedules the 
broadcast of packets in parent agents and specifies a common packet format used for relative localization 
and clock synchronization in both parent and child agents.

\subsection{D-TDMA scheduling}
\begin{figure}
    \centering
    \includegraphics[width=0.9\linewidth]{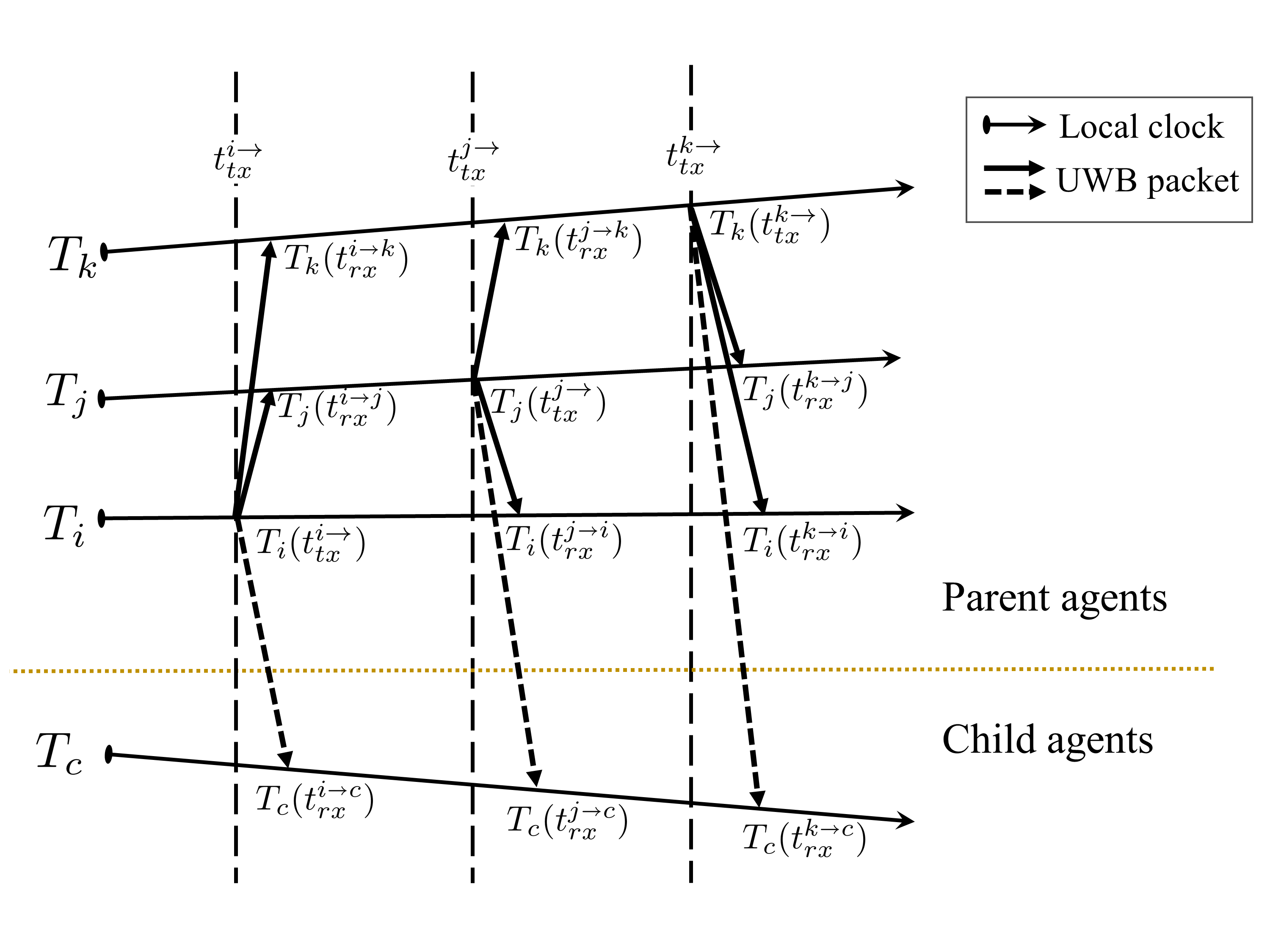} 
    \caption{The ABPR protocol with an example sequence of the broadcast UWB packets, which are scheduled by D-TDMA scheme. 
    Solid lines 
    refer to the packets received by parent agents, in which case a two-way link is formed. 
    Dotted lines refer to the passively received packets in child agents.}
    \label{fig:protocol2}
\end{figure}

\begin{figure}
    \centering
    \includegraphics[width=0.9\linewidth]{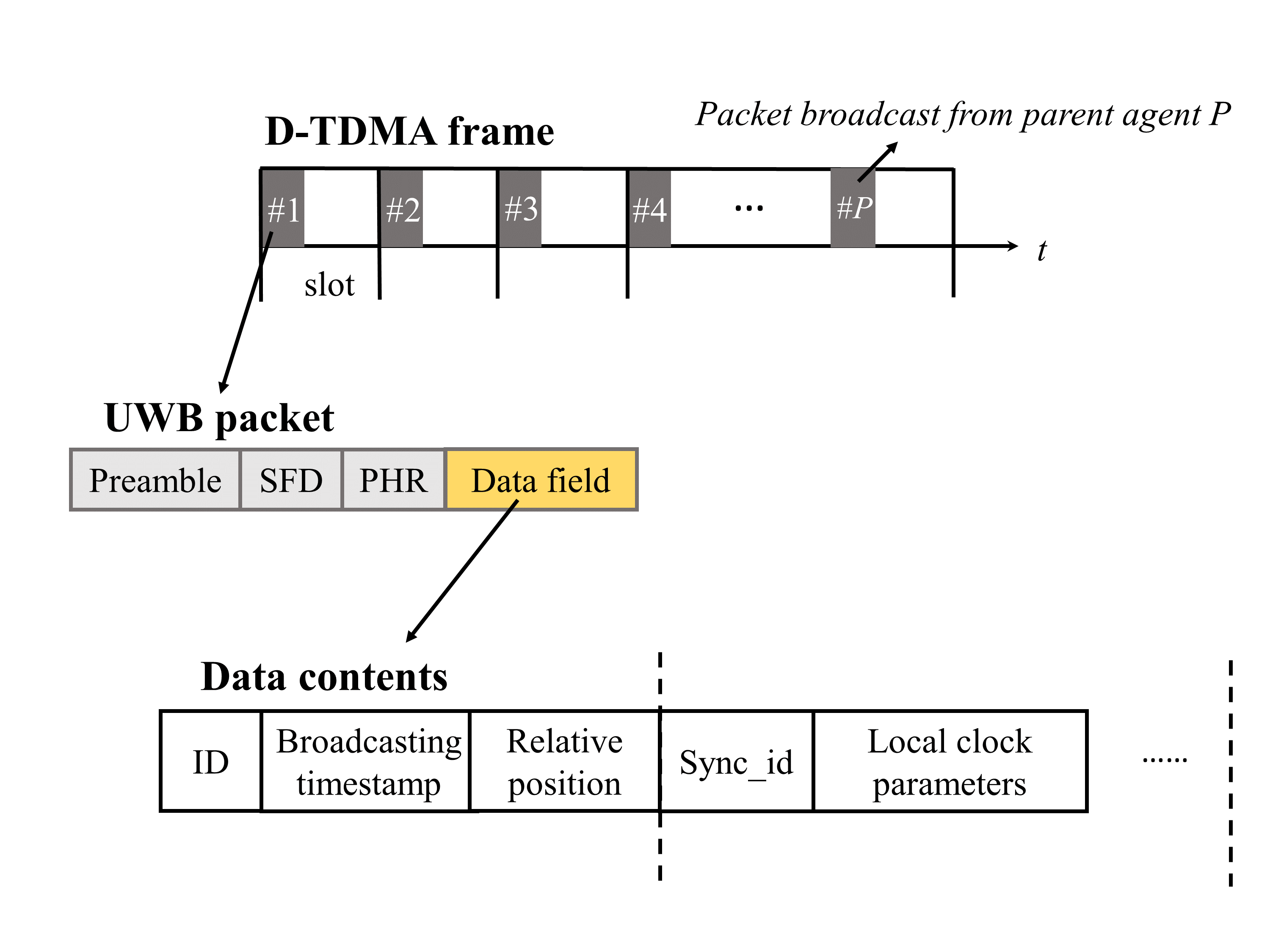} 
    \caption{The ABPR protocol with a D-TDMA frame 
    containing $P$ packets (upper). 
    Each packet contains specially designed data contents for common use (bottom).}
    \label{fig:protocol}
\end{figure}

In order to achieve collision-free broadcasting of UWB packets between parent agents,
the broadcasting is scheduled in a round-robin way as illustrated 
in \mbox{Fig. \ref{fig:protocol2}}.
We employ a distributed time division multiple access (D-TDMA) scheduling scheme,
which is implemented on individual parent agents to decide if and when to broadcast a packet. 
\mbox{Fig. \ref{fig:protocol}} shows a D-TDMA frame containing $P$ UWB packets. 
Typically, agents are allocated a predefined time slot for packet broadcasting \cite{hamer2018self}.
However, the clocks between agents will drift in a long-running, thus leading to a collision
of the predefined time slots. 
Consequently, it is necessary for the MAS to adjust the broadcasting time slots 
during operation.  
This is enabled through information exchange between agents in D-TDMA, which is 
summarized in Algorithm 1.

D-TDMA starts with an initialization procedure. 
Once powered on, parent agent $i$ will listen on the on-air packets for a specific while 
to see if any other parent agent already broadcasting.  
If no packets are received, it is the first one powered on \mbox{(line 3)}. It then
broadcasts a first packet \mbox{(line 4)} and determine the delay time $t_{delay}$
until next packet broadcasting \mbox{(line 5)}. Straightforwardly, the delay time 
is determined by the multiplication of the number of parent agents and time slot 
intervals $\Delta t_s$, i.e., the time interval of a D-TDMA frame.
Otherwise, it will receive a first packet from neighboring parent agents, and extract 
the identification of the broadcasting agent embedded in the packet \mbox{(line 6)}.
It then schedules its broadcasting according to the identifications \mbox{(line 7-10)}.
After initialization, 
subsequent adjustment procedure of D-TDMA adjusts the scheduled broadcasting delay 
upon receiving a packet. 
When parent agent $i$ broadcasts a packet according to scheduling, it immediately sets 
the broadcasting delay of the next packet \mbox{(line 16)}. The broadcasting delay
is then adjusted by comparing the broadcasting agent identifications \mbox{(line 18-22)}. 
Note that, the adjustment is performed repeatedly on every packet reception \mbox{(line 17)}. 
In this way, the broadcasting time slots of parent agents are allocated and adjusted in a 
global scope during run time. 

Since the D-TDMA is performed in real time, parent agents may be 
inserted to or removed from the MAS during operation, making the 
system dynamically scalable and robust to failure. 
For example, when a new parent agent is inserted, 
it will first listen to the on-air UWB signals and 
then synchronize to the network to find an available 
time slot. If a time slot is available, it successfully 
joins the network and then begins to broadcast UWB packets. 
If no time slots are available, we can theoretically combine 
other channel access methods, such as frequency division multiple 
access (FDMA) to expand the capability of the UWB network.

\begin{algorithm}  
    \caption{D-TDMA Scheduling}
    \begin{algorithmic}[1]
        \State let $i$ be an arbitrary parent agent.
        \Procedure{Initialization}{}
            \If {$i$ is the first one powered on}
                \State  Broadcast a first packet. 
                \State $t_{delay} \leftarrow P\cdot\Delta t_s$.
            \ElsIf{Receive a first packet from $j\in \mathcal{P}\setminus \{i\}$}
                \If{$j<i$}
                \State $t_{delay} \leftarrow (i-j)\cdot \Delta t_s$.
                \ElsIf{$j>i$}
                \State $t_{delay} \leftarrow (i+P-j)\cdot \Delta t_s$.
                \EndIf
            \EndIf
        \EndProcedure
        \Procedure{Adjustment}{}
            \State Broadcast a packet according to scheduling.
            \State $t_{delay} \leftarrow P\cdot\Delta t_s$.
            \If{Receive a packet from $j\in \mathcal{P}\setminus \{i\}$}
                \If{$j<i$}
                \State $t_{delay} \leftarrow (i-j)\cdot \Delta t_s$.
                \ElsIf{$j>i$}
                \State $t_{delay} \leftarrow (i+P-j)\cdot \Delta t_s$.
                \EndIf
            \EndIf
        \EndProcedure
    \end{algorithmic}
\end{algorithm}

\subsection{Packet format}
D-TDMA requires information of the broadcasting agent identification embedded in 
the packet. Furthermore, clock synchronization and relative localization also 
require information exchange between agents.
To this end, we specify the packet format in \mbox{Fig. \ref{fig:protocol}}.
Each packet begins with a synchronization header consisting of ranging preamble 
and a start-of-frame delimiter (SFD), after which follow the PHY 
header (PHR) and a data field. The packet format is compatible with the IEEE 802.15.4a 
standard \cite{karapistoli2010overview}.
Data contents are embedded in the packet data field, which contains the unique 
identification index $i \in \mathcal P$, the reported broadcasting
timestamp $T_i(t_{tx}^{i\shortrightarrow})$, the relative position $\mathbf x_i$,
and local clock parameters $\mathbf C_i$ evaluated at the broadcasting time. 
The local clock parameters are described in Section V-A.

We note that the relative position and local clock parameters are tracked 
between parent agents using the reported broadcasting timestamp,
as described in Section V, 
and the states of child agents are determined using all the data contents, 
as described in Section VI. 
To this end, the specially designed packet format can be used for general purposes,
which significantly reduces the communication overhead.

Given specific D-TDMA slot interval, the system broadcast rate is 
dependent on the number of parent agents, 
as the child agents only receive packets.
In practice, we set the slot interval $\Delta t_s=1$ ms, which is longer than the 
UWB packet length of approximate 0.3 ms.
By deploying $P=10$ 
parent agents, we achieve a 100 Hz broadcast rate. This enables a relatively 
high-frequency estimation of clock parameters and relative positions, which supports
MASs with high maneuverability.

\section{JLAS of Parent agents}
This section illustrates how parent agents can jointly track their relative positions 
and clock parameters using two-way broadcast packets.
The JLAS is conducted in a two-step approach, which first 
performs clock synchronization and then relative localization. 
In contrast to \cite{denis2006joint}, which adopts a similar two-step approach 
for static networks with low updating rate, we focus on the high-frequency 
and real-time estimation for parent agents.

We choose to track the local clock parameters for every parent agent using 
two-way TOA measurements.
Further, we do not physically modify the internal clocks and just track the 
clock parameters to achieve virtual clock synchronization. 
This is in contrast to the standard clock synchronization approaches, which 
aim to achieve optimal global clock parameters and synchronize every clock 
to this optimal global reference. In practice, we 
implement the clock synchronization algorithms in a distributed way.
Every parent agent locally holds multiple Kalman filters to track the pairwise 
pseudo-clock parameters with respect to other parent agents. 
This enables high-frequency and real-time clock parameter estimation. 
After two-way synchronization completion, the two-way range between a 
pair of parent agents 
can be derived in one round, i.e., in one D-TDMA frame. 
This is time-efficient when compared to conventional two-way ranging 
algorithms\cite{jiang2007asymmetric, neirynck2016alternative}, 
which require two rounds due to unavailability of clock skew 
information. 
We assume that in one D-TDMA frame length, which is a relatively short time, 
e.g., 10 ms,
the motion of parent agents is negligible and
the range between parent agents remains constant. 
We can then recover the topology of parent agents from inter-agent ranges
with 100 Hz frequency.

\subsection{Distributed clock synchronization in parent agents}
Since the parent agents are dynamic, the propagation delay $\tau$ is not
deterministic. As we can see from (\ref{eq:TOA model}), the relative clock parameters
are then not observable using only one-way TOA measurements. 
Therefore, we propose to track the pseudo-clock parameters which consist of the 
relative clock parameters and the propagation delay in the parent agent's local 
perspective.

Without loss of generality, we discuss the case of an arbitrary parent agent $i\in \mathcal P$ 
synchronizing to its neighboring parent agents $\mathcal{P}\setminus \{i\}$. 
The local clock parameters to be estimated are defined as 
$\mathbf C_i=[\mathbf C_i^1, \cdots, \mathbf C_i^P]$, and 
$\mathbf C_i^j$ are the pseudo-clock parameters of $i$ with respect to $j$:
$$
    \mathbf C_i^j = [\widetilde{T}_i^j,\omega_i^j],
$$
where 
\begin{equation} \label{eq:pseudo-clock offset}
    \widetilde{T}_i^j=T_i^j+ \tau_i^j,
\end{equation}
is the pseudo-clock offset and $\omega_i^j$ is clock 
skew of $i$.
For each $\mathbf C_i^j$, we 
establish a Kalman filter. 
Under the assumption that the motion of parent agents is negligible in a small interval 
(10 ms), the Kalman filter linear dynamic model is given using (\ref{eq:relative clk model}):
$$
    \dot{\mathbf C_i^j}(t) = \mathbf F \mathbf C_i^j(t) + \mathbf n(t).
$$
Further, the states are observed upon the reception of an UWB packet by (\ref{eq:TOA model}),
with the Jacobian matrix $\mathbf H$ defined with respect to the state as
$\mathbf H = 
\begin{bmatrix}
    1 & 0 
\end{bmatrix}$.

\begin{algorithm}  
    \caption{Distributed clock synchronization in parents}
    \begin{algorithmic}[1]
        \State let $i$ be an arbitrary parent agent.
        \For {$j\in \mathcal{P}\setminus \{i\}$}
            \State Initialize $j$th Kalman filter with initial state $\mathbf C^{j}_{i,0}$ 
            and initial state error covariance matrix $\mathbf P^{j}_{i,0}$.
        \EndFor
        \While {system running}
            \State Receive the $k$th packet from agent $j\in \mathcal{P}\setminus \{i\}$.
            \State Extract the TOA measurement: (\ref{eq:TOA meas}).
            \State Calculate the reception interval:
            $\Delta t=T^{j\shortrightarrow i}_{rx,k} - T^{j\shortrightarrow i}_{rx,k-1}$.
            \State Compute the discrete-time state transition matrix:
            $$
                \mathbf F_d = \mathbf I_2 + \Delta t \mathbf F.
            $$
            \State Propagate the $j$th Kalman filter:
            $$
                \mathbf C^{j}_{i,k} \leftarrow \mathbf F_d\mathbf C^{j}_{i,k-1}, 
            $$
            $$
                \mathbf P^{j}_{i,k} \leftarrow \mathbf F_d\mathbf P^{j}_{i,k-1}
                    \mathbf F_d^T + \mathbf Q.
            $$
            \State Update the $j$th Kalman filter using TOA $\tau_k^{j\shortrightarrow i}$:
            $$
                \mathbf K = \mathbf P^{j}_{i,k} \mathbf H^T
                    (\mathbf H\mathbf P^{j}_{i,k}\mathbf H^T + \xi^2)^{-1},
            $$
            $$
                \mathbf C^{j}_{i,k} \leftarrow \mathbf C^{j}_{i,k} + 
                    \mathbf K(\tau_k^{j\shortrightarrow i} - \widetilde{T}_{i,k}^j),
            $$
            $$
                \mathbf P^{j}_{i,k} \leftarrow (\mathbf I_2 - \mathbf K\mathbf H)\mathbf P^{j}_{i,k}.
            $$
        \EndWhile
    \end{algorithmic}
\end{algorithm}

Algorithm 2 outlines the recursive state estimation steps for parent agent 
synchronization using discrete-time Kalman filters. 
This algorithm is implemented in each parent agent.
During initialization, a number of $P-1$ Kalman filters are established (line1-3).
Upon reception of an UWB packet, the parent agent reads the broadcasting agent id
and extract the TOA measurement (line 6-7).
Then it switches into the corresponding Kalman filter process and calculates the 
discrete reception time interval (line 8).
The Kalman filter state and covariance matrix is then propagated (line 10)
and updated using the TOA measurement (line 11), where the discrete-time 
noise covariance matrix is given by \cite{brown1992introduction}
$$
\mathbf Q = 
\begin{bmatrix}
    2S_{n_T}\Delta t + 2S_{n_\omega}\frac{\Delta t^3}{3} & S_{n_\omega}\Delta t^2 \\
    S_{n_\omega}\Delta t^2 & S_{n_\omega} \Delta t
\end{bmatrix}.
$$

During implementation, we also set thresholds for the values of the estimation error covariance 
matrix to detect whether a parent agent have left the network. 
The covariance grows if no UWB packets are received from the left parent agents,
and if it exceeds the thresholds, the associated clock parameters will not be embedded in the 
broadcast packet.

\subsection{Inter-agent range estimation in parent agents}
\begin{figure}  
    \centering
    \includegraphics[width=0.9\linewidth]{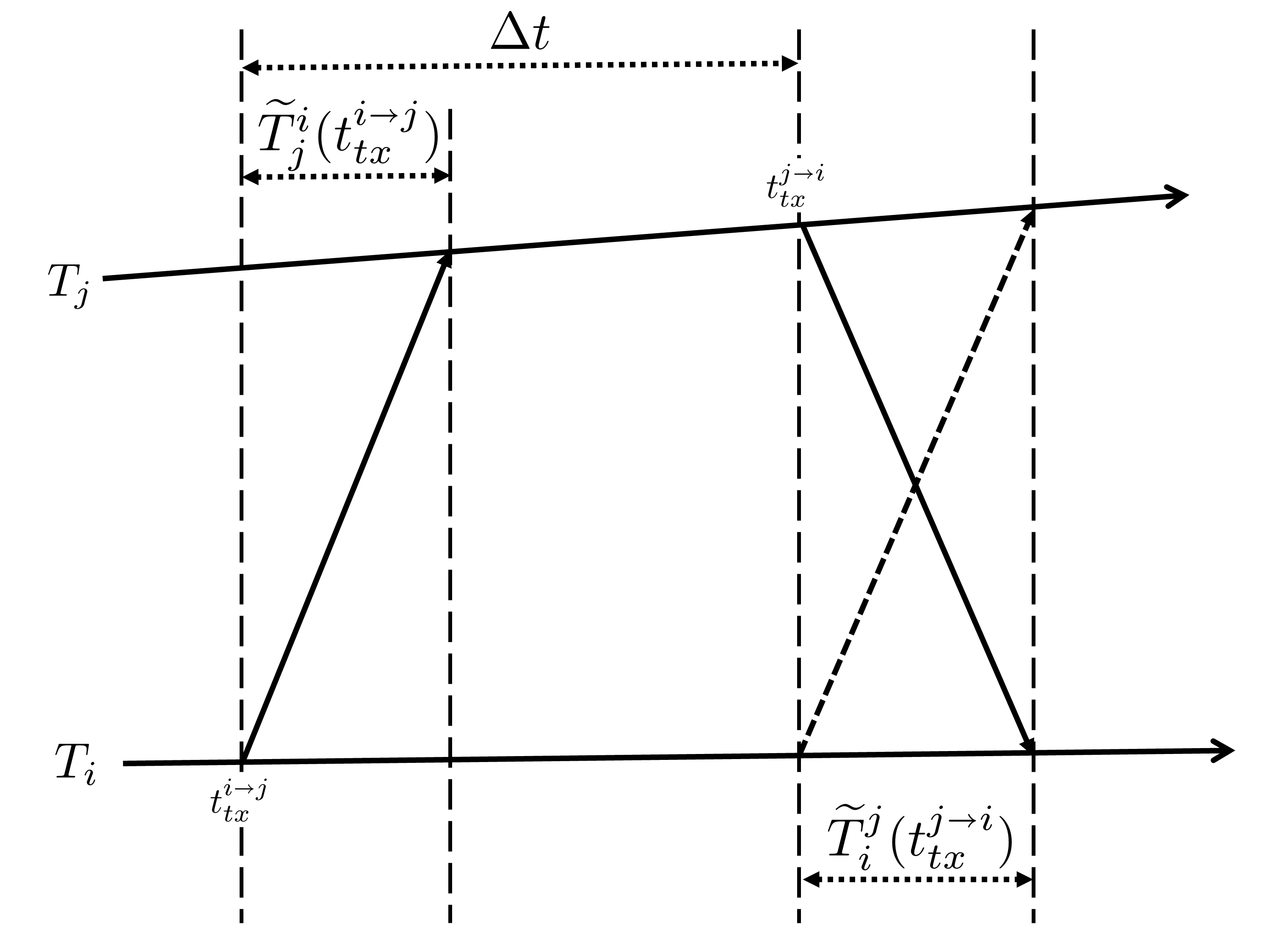} 
    \caption{Packet sequence in two-way ranging. Parent agent $i$ first broadcasts an UWB packet and then receives one 
    from parent agent $j$ (solid lines). However, two-way ranging requires pseudo-clock parameters 
    $\widetilde{T}_j^i(t^{j\shortrightarrow i}_{tx})$ and $\widetilde{T}_i^j(t^{j\shortrightarrow i}_{tx})$ 
    estimated at the same time $t^{j\shortrightarrow i}_{tx}$, which means that parent agent $i$ should physically 
    broadcast a packet at $t^{j\shortrightarrow i}_{tx}$ (the dotted line). This is unpractical. Therefore, we choose to propagate the pseudo-clock parameters
    in agent $j$ until broadcasting as an approximation of $\widetilde{T}_j^i(t^{j\shortrightarrow i}_{tx})$.
    }
    \label{fig:twr} 
\end{figure}

When the local clock parameters are acquired in parent agents, the inter-agent 
range can be estimated from the exchanging packets. 
Consider a pair of parent agents $i$ and $j,(i<j)$ as illustrated 
in \mbox{Fig. \ref{fig:twr}},
parent agent $i$ first broadcasts an UWB packet and then receives one 
from parent agent $j$ in one D-TDMA frame interval.
The relative clock offset of $i$ with respect to $j$ 
at broadcasting time $t^{j\shortrightarrow i}_{tx}$,
$T_i^j(t^{j\shortrightarrow i}_{tx})$  
is given by (\ref{eq:clock offset}):
$$T_i^j(t^{j\shortrightarrow i}_{tx}) = 
    T_i(t^{j\shortrightarrow i}_{tx}) - T_j(t^{j\shortrightarrow i}_{tx}) \\
    =-T_j^i(t^{j\shortrightarrow i}_{tx}),
$$
which is the negative of relative clock offset $T_j^i(t^{j\shortrightarrow i}_{tx})$ 
of $j$ with respect to $i$ at the same time.
Using (\ref{eq:pseudo-clock offset}), 
$\tau_i^j$ can be written in terms of the pseudo-clock offset estimated respectively in 
$i$ and $j$ as: 
\begin{equation} \label{eq:signal delay est}
    \tau_i^j(t^{j\shortrightarrow i}_{tx}) = \frac{1}{2}(\widetilde{T}_j^i(t^{j\shortrightarrow i}_{tx}) + 
        \widetilde{T}_i^j(t^{j\shortrightarrow i}_{tx})).
\end{equation}

However, this requires parent agent $i$ and $j$ broadcast a packet at the same time to 
estimate the corresponding pseudo-clock parameters and exchange them without
latency, which is unpractical for D-TDMA scheduling scheme.
Therefore, we choose to propagate the pseudo-clock parameters
in agent $j$ until broadcasting as an approximation of $\widetilde{T}_j^i(t^{j\shortrightarrow i}_{tx})$:
$$
    \widetilde{T}_j^i(t^{j\shortrightarrow i}_{tx}) \cong 
        \widetilde{T}_j^i(t^{i\shortrightarrow j}_{tx}) + \Delta t \cdot 
        \omega_j^i(t^{i\shortrightarrow j}_{tx}),
$$
where $\Delta t = T_j(t^{i\shortrightarrow j}_{tx}) -  T_j(t^{j\shortrightarrow i}_{tx}) \approx (j-i)\cdot \Delta t_s$.
This is the main reason why 
we embed the propagated pseudo-clock parameters in the UWB packets as the aforementioned 
protocol. 

To this end, after reception of an UWB packet from $j$, the parent agent $i$ can 
estimate the pseudo-clock offset $\widetilde{T}_j^i(t^{j\shortrightarrow i}_{tx})$ and 
extract the embedded pseudo-clock offset $\widetilde{T}_i^j(t^{j\shortrightarrow i}_{tx})$, 
the inter-agent range can then be estimated using (\ref{eq:signal delay est}):
\begin{equation} \label{eq:twrwithbias}
    d^{ij}(t^{j\shortrightarrow i}_{tx})= v_c \cdot 
    \Big(\frac{\widetilde{T}_j^i(t^{j\shortrightarrow i}_{tx}) + 
    \widetilde{T}_i^j(t^{j\shortrightarrow i}_{tx})}{2} - \tau_i - \tau_j \Big).
\end{equation}

We note that the distance is estimated using two-way packets in one round, i.e., in 
one D-TDMA frame. Even with the case of packet loss, for example, the $k$th packet from $i$ to 
$j$ is lost, parent agent $i$ can still estimate the distance as long as it receives the 
$k$th packet from $j$ to $i$, since we can propagate $\widetilde{T}_j^i(t^{j\shortrightarrow i}_{tx})$
according to $\Delta t$ (though with increasing uncertainty). 
Also we note that the inter-agent ranges are estimated in a distributed way. 
Each parent agent
is able to track the pair-wise distances between parent agents: 
$\{d^{ij} \mid i \in \mathcal P,j \in \mathcal P, i \neq j\}$.
Since they are aware of the pair-wise clock parameters by receiving the broadcast UWB packets.
Consequently, a distributed relative localization 
algorithm in parent agents can be achieved.

\subsection{Relative localization in parent agents}
If the inter-agent ranges are known for each pair of parent agents, the topology 
of parent agents can be recovered. In other words, the relative positions of each parent agents 
can be estimated. 
In the context of WSN localization, given the inter-agent range measurements, 
a configuration of the sensor positions can be estimated using the classical MDS method.
A distributed implementation of MDS can be found in \cite{costa2006distributed}.
However, high-maneuverability MASs are with varying topology, 
thus requiring real-time and high-frequency relative localization. Further, SWaP constrained
agents pose additional requirement on the complexity of localization algorithms. 
Therefore, in the context of dynamic MASs, it is often hard to adapt the MDS method, 
which requires many computation resources to minimize a rather complex cost function. 

To this end, we choose to determine the topology by assuming that the parent agents $1,2,3$
can uniquely construct a 2D reference Cartesian coordinate. The origin of the coordinate is set as the 
position of agent $1$, positive x-axis along the parent agent $2$, and positive y-axis in the direction of 
parent agent $3$. The rest of parent agents can then determine 
their positions relative to this coordinate using the range measurements with respect to 
the first three parent agents and the broadcast information of parent agents' positions.
In this way, the initial relative position of parent agents can be given in closed-form as:
\begin{align*}
\mathbf x_1 &= \begin{bmatrix}0 & 0\end{bmatrix}^T, \\
\mathbf x_2 &= \begin{bmatrix}d^{21} & 0\end{bmatrix}^T, \\
\mathbf x_i &= 
    \begin{bmatrix}\frac{{d^{21}}^2+{d^{i1}}^2-{d^{i2}}^2}{2d^{21}} 
        & \pm \sqrt{{d^{i1}}^2 - x_i^2}
    \end{bmatrix}^T,i=3,\cdots, P, 
\end{align*}
where the y coordinate of parent agent 3 is set to be positive and the sign of y-coordinates of 
other parent agents is determined by comparing their distance to parent agent 3.

We then use all the mutual distance mesurements to refine the initial results.
We seek to distributely estimate the relative position $\mathbf x_i$ for parent agent $i$ by minimizing the following 
local cost function:
$$
    S(\mathbf x_i) = \frac 1 2 \sum_{j\in \mathcal{P}\setminus \{i\}} (d^{ij} - \Vert \mathbf x_j - \mathbf x_i\Vert)^2,
$$
where the distance $d^{ij}$ and the position $\mathbf x_j$ of its neighboring parent agents 
are extracted from the received UWB packets. 
The optimization problem is solved using an iteratively least squares method under the 
constraints of $x_1 = 0, y_1 = 0, y_2 = 0$.

We note that the accuracy of the relative position estimate is highly 
related to the inter-agent range accuracy. Since we do not have
access to global position and orientation or the odometry information 
over time, we can only determine a rough topology of parent agents in local perspective.
Future improvements to obtain more precise and smooth relative position of parent agents 
would be possible by utilizing additional sensors, such as cameras and inertial measurement units (IMUs).
Then by fusing with inter-agent range measurements, the parent agents can act like 
moving beacons broadcasting UWB packets with their absolute and smooth positions embedded.
This is beyond the scope of this paper and left as our future work.

\section{JLAS of Child Agents}
Along with the synchronized clocks in parent agents, the spatiotemporal 
reference is then obtained in a high frequency and real-time embedded in the 
broadcast UWB packets.
Upon passively receiving the broadcast packets,
our aim is to estimate the state of the child agents $c\in \mathcal C$:
$$
\mathbf X_c^i=[T_c^i(t_{tx}^{i\shortrightarrow c}), \omega_c^i(t_{tx}^{i\shortrightarrow c}),
\mathbf x_c(t_{tx}^{i\shortrightarrow c})],
$$
which consists of the relative clock parameters with respect to a reference parent 
agent $i$ and the relative position.

We choose to utilize an iterative least squares estimator for joint estimation 
using the TOA measurements. 
However, the joint estimation faces two problems. One lies in the D-TDMA scheduling 
scheme, which leads to asynchronous TOA measurements (see \mbox{Fig. \ref{fig:protocol2}}). 
The other lies in the virtual clock synchronization in parent agents, which 
leads to the TOAs measured with respect to different time references. 

Before diving into the estimator, we first investigate the observation equation. 
Consider one round of UWB broadcasting from parent agents, i.e., one D-TDMA frame,
the TOA measurements at 
child agent $c$ are then given by a sequence: 
$$
    [\tau(t^{1\shortrightarrow c}_{tx}),\tau(t^{2\shortrightarrow c}_{tx}),
    \cdots, \tau(t^{P\shortrightarrow c}_{tx})].
$$
Consider the TOA measurement from parent agents $j$
$$
\tau(t^{j\shortrightarrow c}_{tx}) = T_c^j(t^{j\shortrightarrow c}_{tx})
+ \tau_c^j(t^{j\shortrightarrow c}_{tx})
+ n^{cj}. 
$$
We choose an arbitrary reference parent agent $i$. 
From (\ref{eq:clock offset}), we have:
\begin{align*}
    T_c^j(t_{tx}^{j\shortrightarrow c}) &= 
    T_c^i(t_{tx}^{j\shortrightarrow c}) + T_i^j(t_{tx}^{j\shortrightarrow c}) \\
    &\cong 
    T_c^i(t_{tx}^{i\shortrightarrow c}) + 
    \omega_c^i(t_{tx}^{i\shortrightarrow c}) \cdot \Delta t +
    T_i^j(t_{tx}^{j\shortrightarrow c}),  
\end{align*}
where $\Delta t = T_j(t^{j\shortrightarrow i}_{tx}) - T_j(t^{i\shortrightarrow j}_{tx}) \approx (j-i)\cdot \Delta t_s$.
From (\ref{eq:sig delay}), we have:
$$
    \tau_c^j(t^{j\shortrightarrow c}_{tx}) = \tau_j + \tau_c + 
    \frac{\Vert 
    \mathbf x_j(t_{tx}^{j\shortrightarrow c}) - 
    \mathbf x_c(t_{tx}^{i\shortrightarrow c}) \Vert }{v_c}.
$$
The TOA measurement from parent agent $j$ can then be rewritten as:
\begin{multline} \label{eq:TOA 1}
    \tau(t^{j\shortrightarrow c}_{tx}) =         
    T_c^i(t_{tx}^{i\shortrightarrow c}) + 
    \omega_c^i(t_{tx}^{i\shortrightarrow c}) \cdot \Delta t +
    T_i^j(t_{tx}^{j\shortrightarrow c}) \\ +
    \tau_j + \tau_c + 
    \frac{\Vert 
    \mathbf x_j(t_{tx}^{j\shortrightarrow c}) - 
    \mathbf x_c(t_{tx}^{i\shortrightarrow c}) 
    \Vert }{v_c} + n^{cj}. 
\end{multline}
As we can see, the state of child agent can be observed. 
We note $\Delta t$ corresponds to the aforementioned asynchronous TOA problem, and 
$T_i^j(t_{tx}^{j\shortrightarrow c})$ corresponds to the virtual 
clock synchronization problem. If these items are not carefully considered, 
the estimation results will be incorrect.

We now try to recover the relative clock offset $T_i^j(t_{tx}^{j\shortrightarrow c})$
from the local clock parameters extracted from the consecutive UWB packets.
From (\ref{eq:pseudo-clock offset}), 
we arrive at:
\begin{align*}
    T_i^j(t_{tx}^{j\shortrightarrow c}) &= 
    \frac{\widetilde{T}_i^j(t_{tx}^{j\shortrightarrow c}) - 
        \widetilde{T}_j^i(t_{tx}^{j\shortrightarrow c})}{2} \\
    &\cong
    \frac{(\widetilde{T}_i^j(t_{tx}^{i\shortrightarrow j}) + 
    \omega_i^j(t_{tx}^{j\shortrightarrow i})\cdot \Delta t) - 
    \widetilde{T}_j^i(t_{tx}^{j\shortrightarrow i})
    }{2},
\end{align*}
where the pseudo-clock parameters in the second line equation are embedded in UWB packets.
In practice, $T_i^j(t_{tx}^{j\shortrightarrow c})$ can also be derived from other pairs of 
relative clock parameters:
$$
T_{i,m}^{m,j}(t_{tx}^{j\shortrightarrow c}) = T_i^m(t_{tx}^{j\shortrightarrow c}) + 
T_m^j(t_{tx}^{j\shortrightarrow c}),
$$
where $T_i^m(t_{tx}^{j\shortrightarrow c})$ and $T_m^j(t_{tx}^{j\shortrightarrow c})$
are computed from the corresponding pseudo-relative clock parameters,
$m\in \mathcal P \setminus \{i,j\}$.
In this way, the relative clock offset can be refined by averaging:
\begin{equation}  \label{eq:actual clock offset}
    T_i^j(t_{tx}^{j\shortrightarrow c}) \leftarrow 
\frac{T_i^j(t_{tx}^{j\shortrightarrow c}) + 
\sum\limits_{m\in \mathcal P \setminus \{i,j\}}T_{i,m}^{m,j}(t_{tx}^{j\shortrightarrow c})}{P-1}.
\end{equation}
To the end, we can obtain the pseudo-range observation equation by multiplying light
speed with (\ref{eq:TOA 1}):
\begin{align*}
    h_j(\mathbf X_c^i) =  
    v_c \cdot (T_c^i(t_{tx}^{i\shortrightarrow c}) + 
    \omega_c^i(t_{tx}^{i\shortrightarrow c}) \cdot \Delta t +
    T_i^j(t_{tx}^{j\shortrightarrow c}) \\ +
    \tau_j + \tau_c) + 
    \Vert 
    \mathbf x_j(t_{tx}^{j\shortrightarrow c}) - 
    \mathbf x_c(t_{tx}^{i\shortrightarrow c}) 
    \Vert +   
    v_c \cdot n^{cj}.  
\end{align*}

\begin{algorithm}  
    \caption{JLAS for child agents}
    \begin{algorithmic}[1]
        \State let $c \in \mathcal C$ be an arbitrary child agent.
        \While {Received a D-TDMA UWB frame}
            \State Get TOA measurements sequence 
            \mbox{$\{\tau(t^{j\shortrightarrow c}_{tx}) \mid j \in \mathcal P\}$}: 
            $$
            (\ref{eq:TOA meas}).
            $$
            \State Recover pseudo-range measurements \mbox{$\{\rho_j \mid j \in \mathcal P\}$}: 
            $$
                \rho_j = v_c \cdot \tau(t^{j\shortrightarrow c}_{tx}).
            $$
            \State Extract the broadcast local clock parameters sequence \mbox{$\{\mathbf C_j(t_{tx}^{j\shortrightarrow c})\mid j \in \mathcal P\}$}.
            \State Extract position of parent agents $\{\mathbf x_j \mid j \in \mathcal P\}$.
            \State Get relative clock offset sequence \mbox{$\{T_j^i(t_{tx}^{j\shortrightarrow c})\mid j \in \mathcal P\}$}: 
            $$
            (\ref{eq:actual clock offset}).
            $$
            \State Iteratively solve for $\bar{\mathbf X_c^i}$: 
            $$
                \underset{\mathbf X_c^i}{\mathrm{arg\,min}}\,S(\mathbf X_c^i) := \sum\limits_{j \in \mathcal P}\Vert \rho_j - h_j(\mathbf X_c^i) \Vert^2.
            $$
        \EndWhile
    \end{algorithmic}
\end{algorithm}

Algorithm 3 outlines the JLAS of child agents. Without loss of generality,
we let reference parent agent $i=1$.
By stacking all the pseudo-range measurements, we can form a least squares problem 
and iteratively solve it to jointly estimate the child agent's state.

\section{Simulations}
In this section, we demonstrate the performance of our proposed JLAS algorithm for 
fully dynamic MASs in numerical simulations. The simulation configuration is given 
first, the results are then provided.

\begin{figure}
    \centering 
    \includegraphics[width=\linewidth]{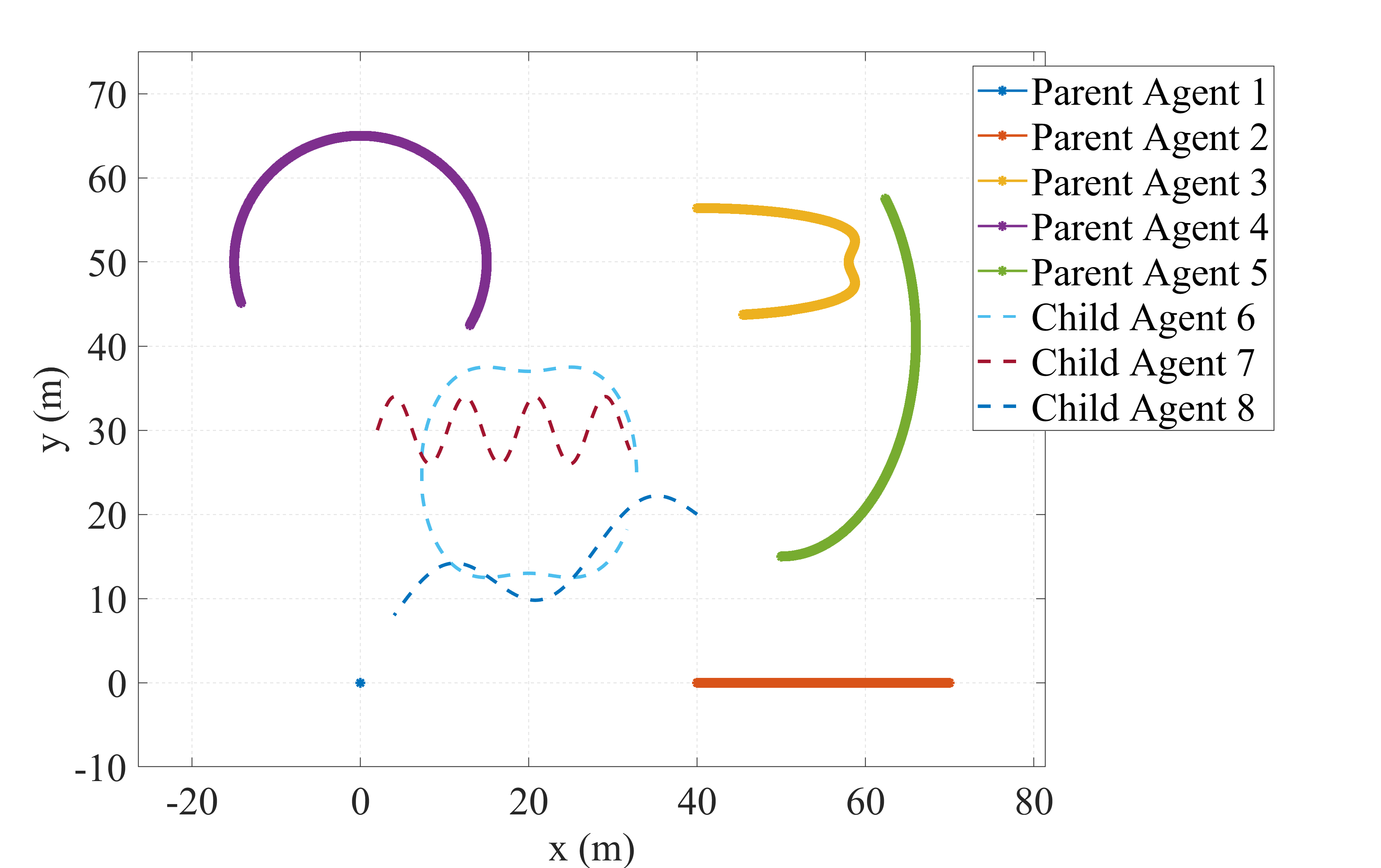} 
    \caption{Simulated trajectories of a MAS with five parent agents and three child agents.}
    \label{fig:simulated_traj}
\end{figure}

\textit{1) Simulation setup:}
The simulated scenario of a MAS consisting of 5 parent agents and 3 child agents, 
labeled with their identification indices as 
$1,\cdots,8$, is illustrated in \mbox{Fig. \ref{fig:simulated_traj}}. 
We preallocate 5 time slots with a slot interval \mbox{$\Delta t_s = 0.001$ s}.
The initial clock offset $t_{i,0}^0$ and initial clock skew $\omega_{i,0}^0$ of each agent $i$ are set 
as random variables, which have continuous uniform distributions, 
\mbox{$t_{i,0}^0 \sim \mathcal U(-5 \times 10^{-7}, 5 \times 10^{-7})$ s} and 
\mbox{$\omega_{i,0}^0 \sim \mathcal U(-5, 5)$ ppm}.
The dynamics of the clock of each agent are then simulated using (\ref{eq:ideal clk model}).
To simplify the comparison of simulation results with ground truth trajectories,
parent agent $1$ is set to be static at \mbox{$(0,0)$ m} and 
parent agent $2$ is set to move along the x-axis, therefore the process of trajectory alignment
can be ignored. 
We note that this setup does not have any implication on the simulations, since we focus on the 
relative localization with respect to the reference coordinate rigidly attached to parent agent $1$.
The eight agents are initially located at $(0,0)$, $(40,0)$, $(40, 56.4)$, $(13,42.5)$,
$(50,15)$, $(32.8, 25)$, $(2,30)$ and \mbox{$(40,20)$ m}. 
The temporal precision of TOA measurements is associated with the noise $n^{ij}$ in 
(\ref{eq:TOA meas}) with standard deviation $\xi$.
We set \mbox{$\xi=1.5 \times 10^{-10}$ s}, which is equivalent to a 
spatial precision of \mbox{0.045 m}. The antenna delays and the channel imperfections, 
such as non-line-of-sight (NLOS) effects, are not considered here.
The initial estimation covariance matrices of the Kalman filters for clock synchronization in parent 
agents are set as $\mathbf P_{i,0}^{j}=\textrm{diag}[0.1, 1]$. The simulation time length is set to 
be \mbox{$60$ s}.
The simulation configurations are summarized 
in \mbox{Table \ref{table:simulation Settings}}. 
 
\begin{table}[!t]
\renewcommand{\arraystretch}{1.5}
\caption{Simulation Settings}
\label{table:simulation Settings}
\centering
\begin{tabular}{ c c c }
\hline \hline
Parameter & Value & Unit \\
\hline
$P$ & 5 & - \\
$C$ & 3 & - \\
$\Delta t_s$ &0.001 & s \\
$S_{n_T}$ & $4.7 \times 10^{-20}$ & s \\
$S_{n_{\omega}}$ & $7.5\times 10^{-20}$ & $s^{-1}$\\
$\xi$ & $1.5 \times 10^{-10}$ & s \\
$t_{i,0}^0$ & $\mathcal U(-5 \times 10^{-7}, 5 \times 10^{-7})$ & s \\ 
$\omega_{i,0}^0$ & $\mathcal U(-5, 5)$ & ppm \\
\hline
\end{tabular}
\end{table}

\begin{figure}
    \centering
    \subfloat[]{\includegraphics[width=\linewidth]{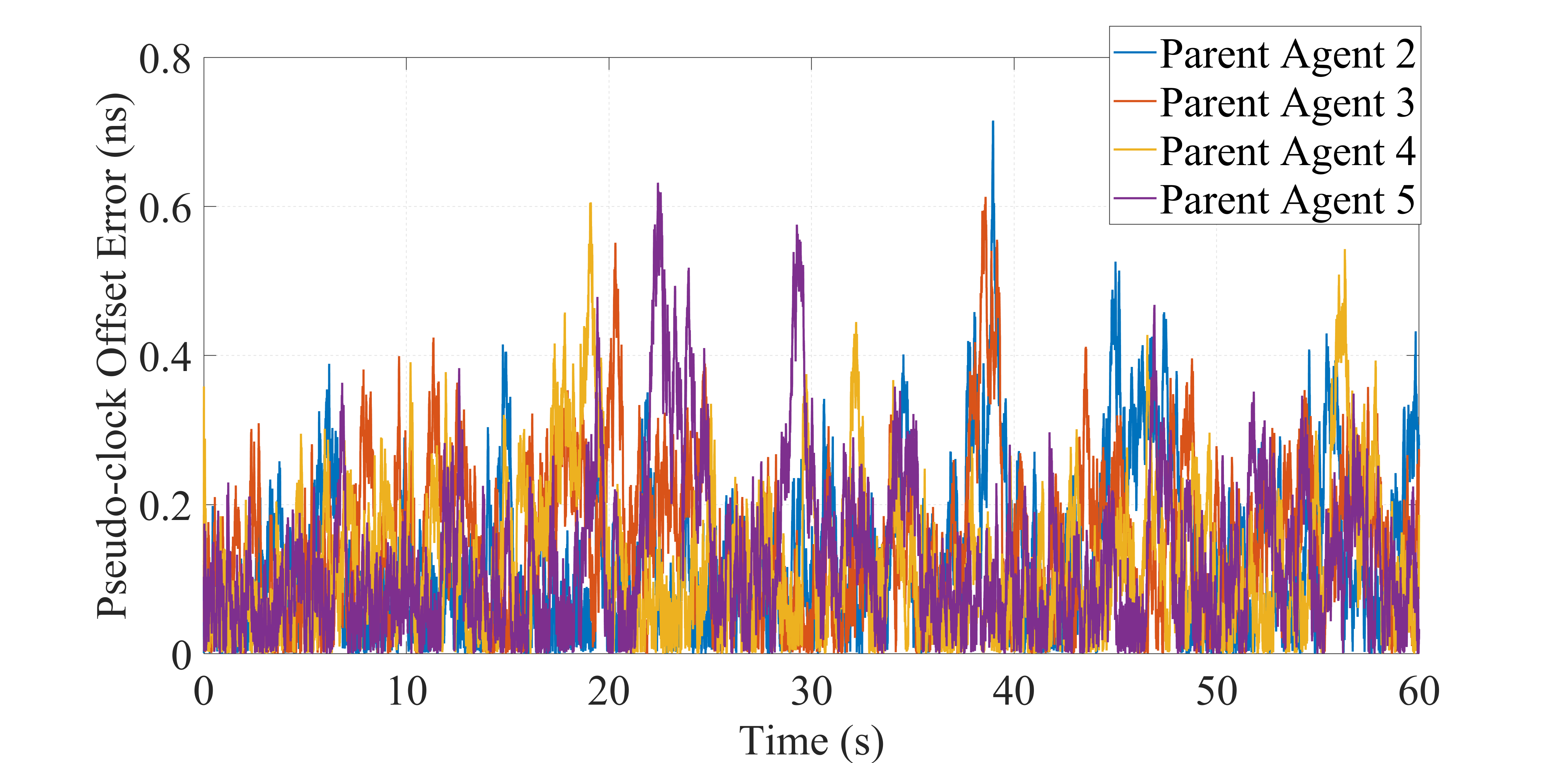}
    \label{fig:sim_pseudo-clock offset error}}
    \hfill
    \subfloat[]{\includegraphics[width=\linewidth]{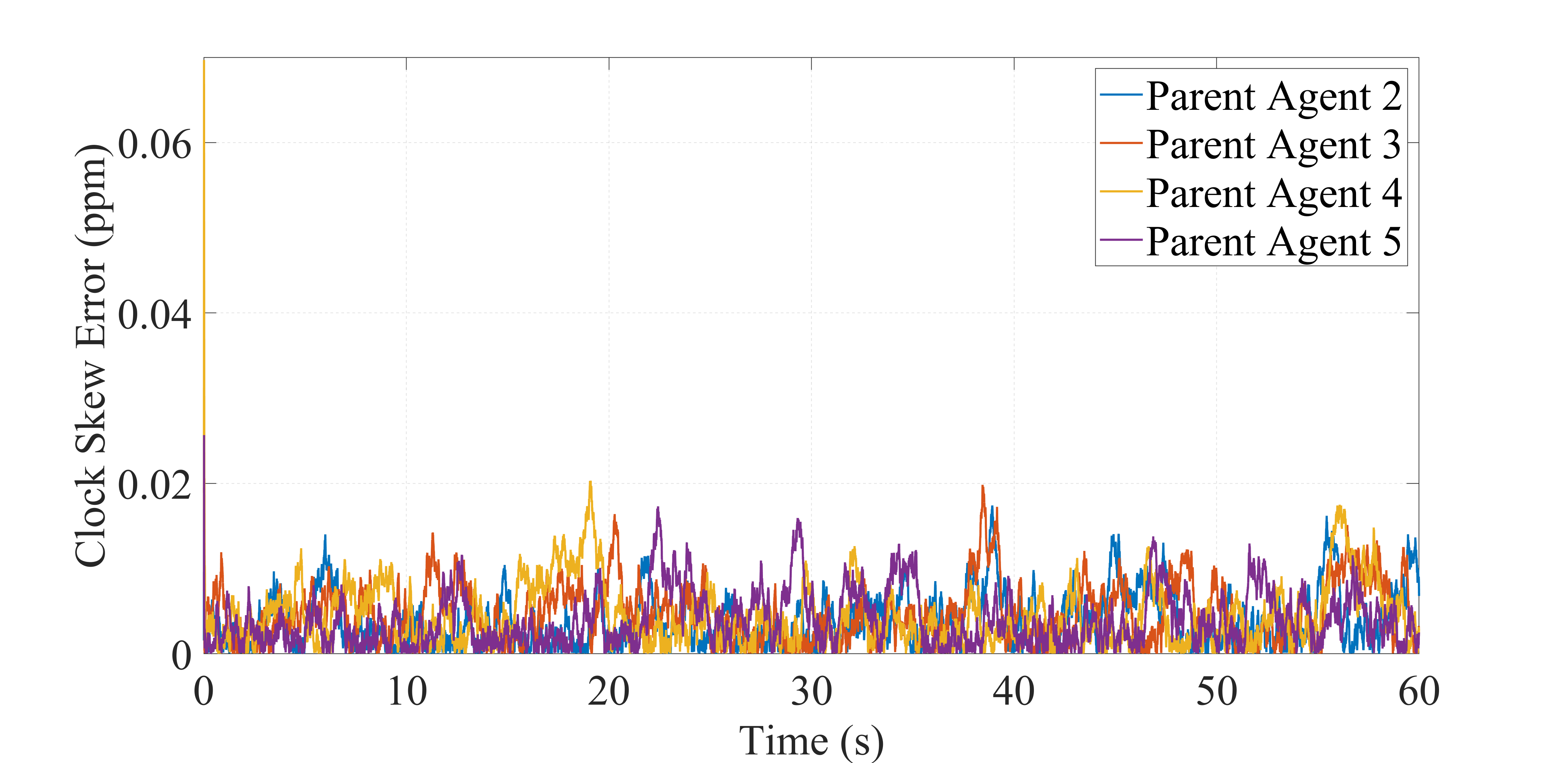}
    \label{fig:sim_clock skew error}}
    \caption{Synchronization results in parent agents. 
    (a) Evolution of pseudo-clock offset error estimated in parent agent $i$, $i=2,\cdots, 5$.
    (b) Evolution of clock skew error estimated in parent agent $i$, $i=2,\cdots, 5$.} 
    \label{fig:sim_parent synchronization}
\end{figure}

\begin{figure}
    \centering
    \subfloat[]{\includegraphics[width=\linewidth]{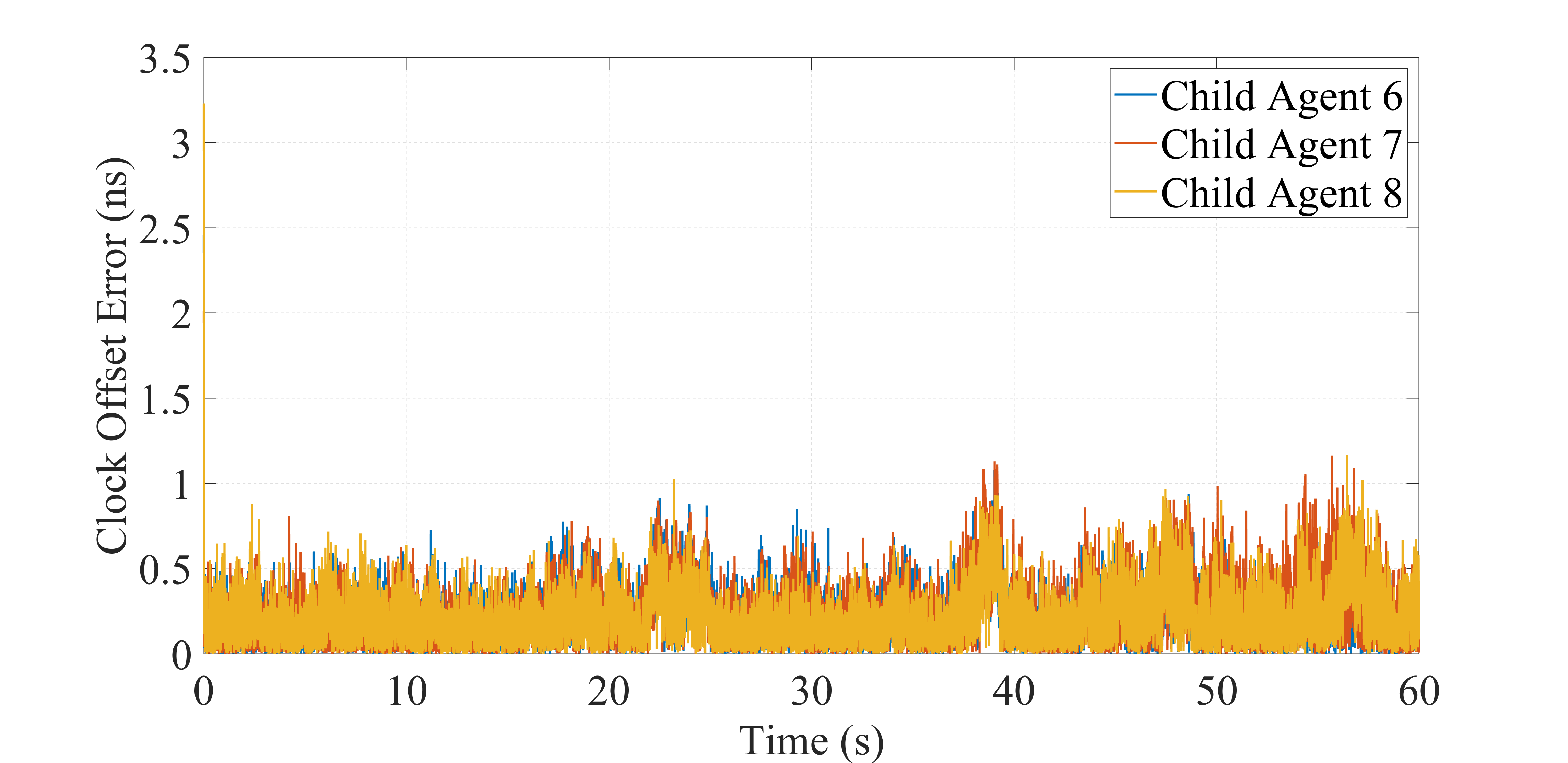}
    \label{fig:sim_child-clock offset error}}
    \hfill
    \subfloat[]{\includegraphics[width=\linewidth]{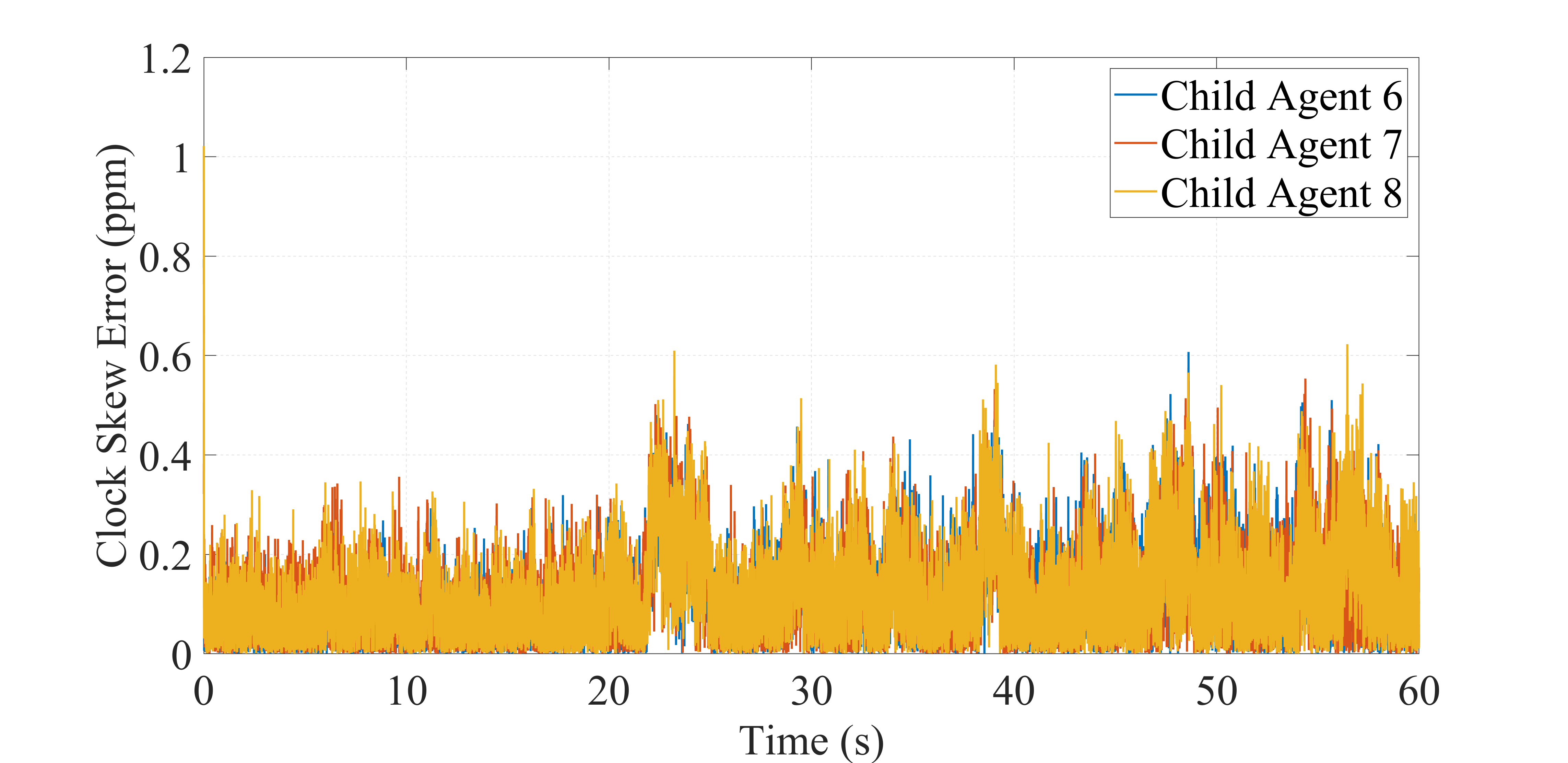}
    \label{fig:sim_child-clock skew error}}
    \caption{Synchronization results in child agents. 
    (a) Evolution of clock offset error estimated in child agent $i$, $i=6,7,8$.
    (b) Evolution of clock skew error estimated in child agent $i$, $i=6,7,8$.} 
    \label{fig:sim_child synchronization}
\end{figure}

\textit{2) Simulation results and discussion:}
We first demonstrate the clock synchronization results and then the relative localization results.
Without loss of generality, we set parent agent $1$ as a reference agent. 
\mbox{Fig. \ref{fig:sim_parent synchronization}} shows the estimation error of pseudo-clock 
parameters $\mathbf C_i^1$ calculated on parent \mbox{agent $i$}, $i = 2,\cdots, 5$.
The statistical results of the clock synchronization errors are presented in \mbox{Table \ref{table:simulation results}}.
The root-mean-square error (RMSE) and standard deviation values of the synchronization errors prove the effectiveness and good performance 
of our proposed algorithm. Note that the synchronization results between other pairs of parent agents, 
namely $\mathbf C_i^j$ for $i\in \mathcal P,j\in\mathcal P, i\neq j$, behave similarly to $\mathbf C_i^1$.
\mbox{Fig. \ref{fig:sim_child synchronization}} illustrates the clock synchronization results in 
child agents. The statistics of the results are summarized in \mbox{Table \ref{table:simulation results}}.
Consider that the actual clock offset can be up to approximately \mbox{$10^{-6}$ s} for the worst case 
according to the simulation configurations in \mbox{Table \ref{table:simulation Settings}},
the average clock offset error of \mbox{$0.21$ ns} thus provides an accuracy of one ten thousandth.
We note that the synchronization results in child agents are worse than those in parent agents.
This comes from the error propagated from the parent agents, since child agents only passively
receive the broadcast information with uncertainty and have no means to observe it.

\begin{figure}
    \centering
    \subfloat[]{\includegraphics[width=\linewidth]{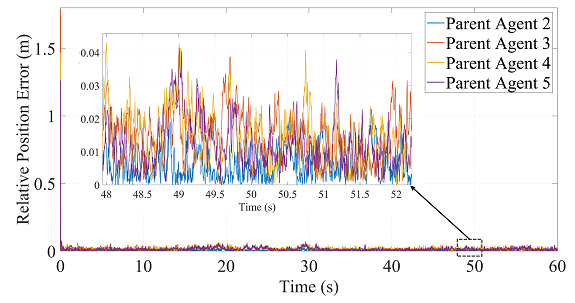}
    \label{fig:sim_parent localization}}
    \hfill
    \subfloat[]{\includegraphics[width=\linewidth]{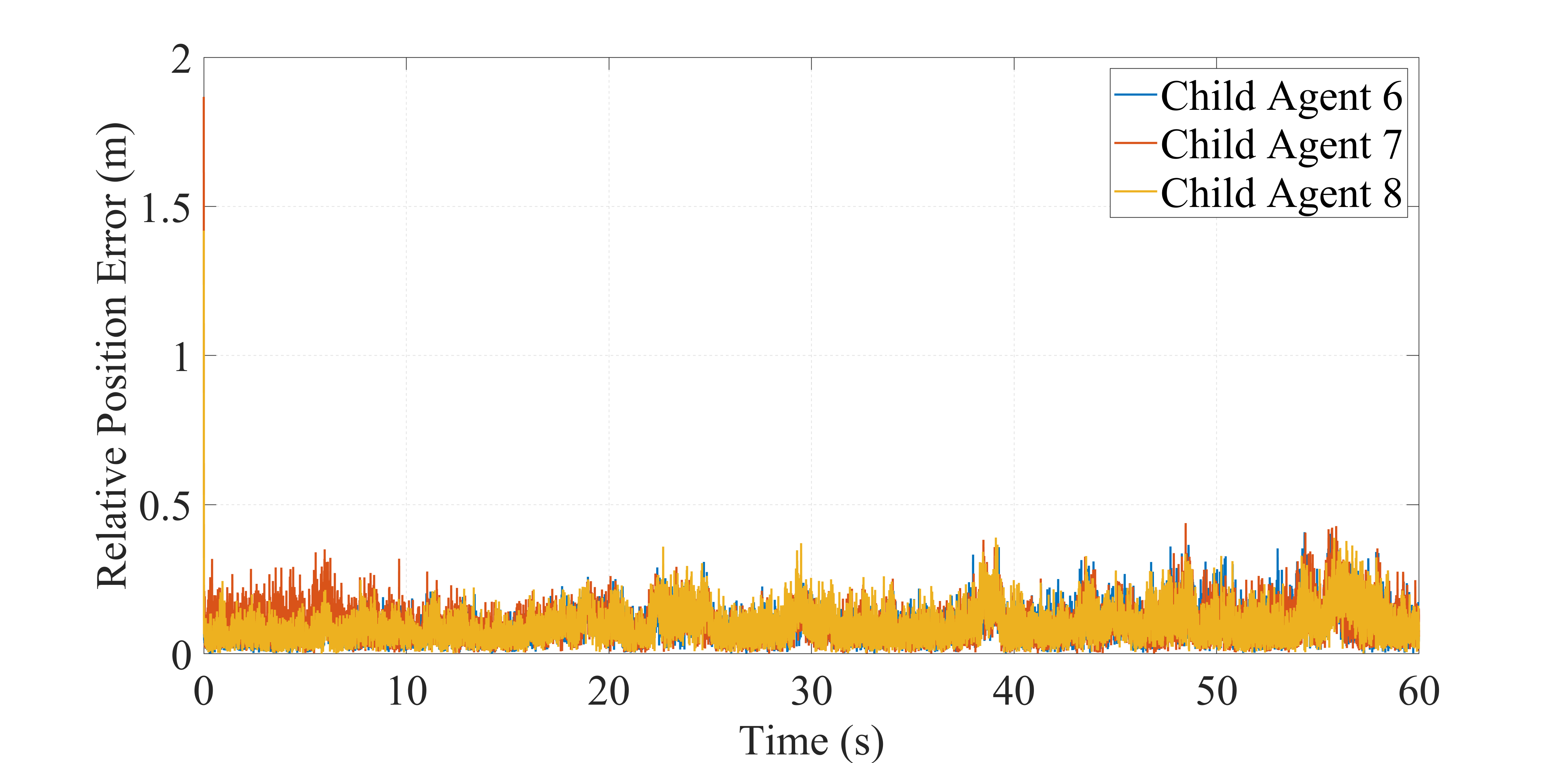}
    \label{sim_child localization}}
    \caption{Relative localization results. 
    (a) Evolution of relative localization error estimated in parent agent $i$, $i=2,\cdots, 5$.
    (b) Evolution of relative localization error estimated in child agent $i$, $i=6,7,8$.} 
    \label{fig:sim_localization}
\end{figure}

We then investigate the relative localization errors. 
The relative localization errors of parent agents and child agents are plotted in \mbox{Fig. \ref{fig:sim_localization}}.
The statistical results are provided in \mbox{Table \ref{table:simulation results}}.
It is easy to see that our proposed method can effectively estimate the relative positions of 
the dynamic agents with estimation errors bounded by approximately \mbox{$0.12$ m}.

Considering 5 time slots with slot interval of \mbox{$\Delta t_s=1$ ms} allocated in our simulations, 
the estimation frequency is then 200 Hz, which is sufficient for collaboration or data fusion 
of a MAS in most scenarios. 
We also note that the agents move at a typical average speed of \mbox{1 m/s} and the number of 
child agents can be theoretically unlimited. 
To summarize, our BLAS can effectively solve the JLAS problem for a fully dynamic MAS with 
high density.

Finally, we would like to discuss the convergence and potential degradation 
of our BLAS algorithms in the presence of antenna delays and NLOS signals.
We first discuss the effect of antenna delays. For parent agents, they will not 
prevent the convergence of the clock synchronization algorithms, since they are deterministic 
parameters. If they are not considered and not calibrated, they will be treated as additional 
pseudo-clock offsets between parent agents. This will pose a significant bias in inter-agent 
range estimation, thus the relative localization algorithm will diverge. 
For child agents, they will not prevent the convergence of the JLAS. As we have mentioned before, 
the delays serve as pseudo-clock offsets, thus only leading to an estimation bias on the relative clock 
offset of the child agent. 
In our practical implementation, the antenna delays are pre-calibrated and compensated. 
We then discuss the effects of NLOS. NLOS will prevent the convergence of the JLAS both in 
parent and child agents. NLOS can be considered as observation outlier as it diverges from 
measurement model assumptions, which introduces a positive bias in range that leads to divergence 
of the algorithms. We note that in our dense MAS scenario, NLOS between agents typically occurs 
due to the occlusions of other agents. A MAS with controlled maneuverability has the potential 
to avoid these NLOS situations. Additionally, since we have a high estimation frequency, fusion 
with additional sensors such as IMUs will give access to NLOS identification and mitigation. 
NLOS avoidance, identification and mitigation is another broad research topic which is out of 
the scope of this paper and left as our future work.

\begin{table}[!t]
    \renewcommand{\arraystretch}{1.6}
    \newcolumntype{Y}{>{\centering\arraybackslash}X}
    \caption{Statistics of JLAS errors}
    \label{table:simulation results}
    \centering
    \begin{tabularx}{\linewidth}{@{}lYYYY@{}}
    \hline \hline
    Agent & (Pseudo-)clock offset (ns) &Clock skew ($1\times 10^{-3}$ ppm) & Relative position (cm)\\
     & \multicolumn{3}{c}{[RMSE, standard deviation]} \\
    \hline
    $2$ & $[0.16, 0.11]$  & \mbox{$[5.0, 3.2]$} & $[0.9, 0.6]$ \\
    $3$ & $[0.18, 0.10]$  & \mbox{$[5.8, 3.4]$} & $[2.7, 2.1]$ \\
    $4$ & $[0.16, 0.09]$  & \mbox{$[5.8, 3.6]$} & $[2.4, 1.6]$  \\
    $5$ & $[0.17, 0.11]$  & \mbox{$[5.1, 3.2]$} & $[2.1, 1.5]$ \\
    \hline
    $6$ & $[0.26, 0.09]$  & $[139.9, 87.9]$ & $[10.8, 5.7]$ \\
    $7$ & $[0.29, 0.18]$  & $[144.2, 89.2]$ & $[11.4, 6.1]$ \\
    $6$ & $[0.27, 0.17]$  & $[151.5, 94.4]$ & $[10.9, 5.8]$ \\
    \hline
    \end{tabularx}
\end{table}

\section{Evaluation}
In this section, we perform two experiments to evaluate the proposed BLAS system. In the first experiment, we evaluate the performance of the proposed algorithm 
for dynamic parent agents. After the spatiotemporal reference is determined by parent agents, we then 
perform an experiment to illustrate that our system supports theoretically unlimited child agents. 
We also show that child agents are able to accurately localize themselves in real-time and high-frequency.

\subsection{Implementation}
\begin{figure}
    \centering
    \includegraphics[width=2.0in]{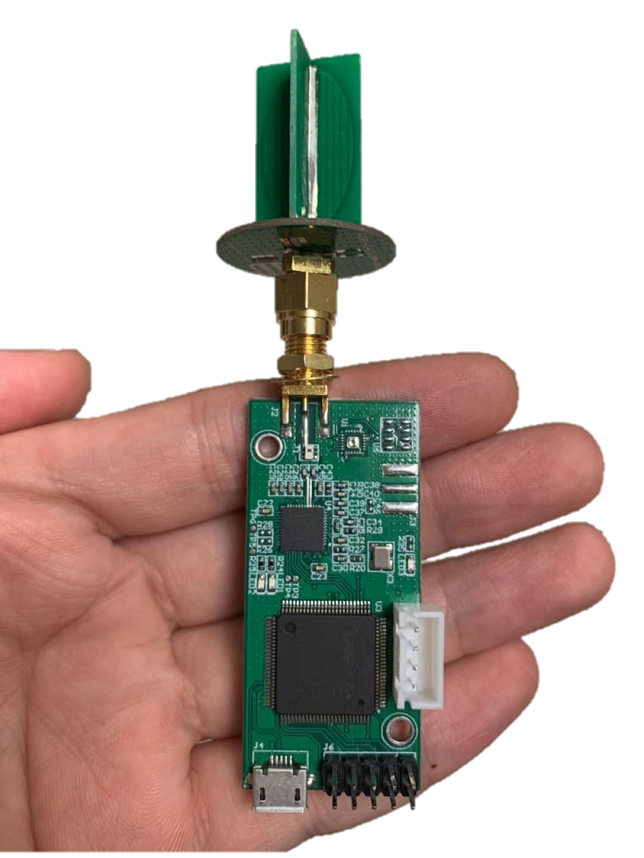} 
    \caption{Self-developed UWB hardware setup which are mounted on parent agents and child agents.}
    \label{fig:hardware}
\end{figure}

We implement our algorithms on the STM32F427 ARM chip, which controls a DecaWave DW1000 UWB module.
\mbox{Fig. \ref{fig:hardware}} shows the hardware setup used in our experiments. Though we categorize 
a MAS into parent and child agents, they are equipped with the same hardware setup and operate 
as a homogeneous team. In this way, a MAS can quickly respond to the failure of a single agent 
by replacing it with a homogeneous one. Further, our setup requires little payload capability and 
onboard resources.
The D-TDMA frame has a length of 10 ms and contains
10 time slots allocated for ten parent agents. This enables 100 Hz UWB packet broadcasting
and thus a 100 Hz estimation frequency.

\subsection{Parent agent experiments}
As we have no ground truth for the clock parameters, we choose to 
evaluate the quality of the ranging results due 
to the coupling between clock synchronization and ranging.
We ignore the evaluation of relative localization results since accurate ranging 
results yield accurate localization results in our relative localization 
algorithm. 

\begin{figure}
    \centering
    \subfloat[]{\includegraphics[width=0.9\linewidth]{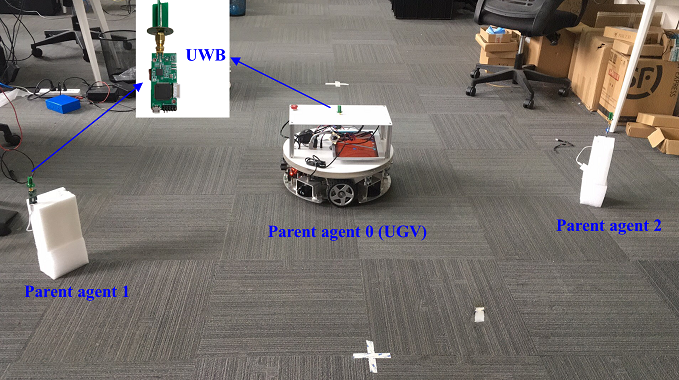}
    \label{fig:exp1_setup1}}
    \hfill
    \subfloat[]{\includegraphics[width=0.9\linewidth]{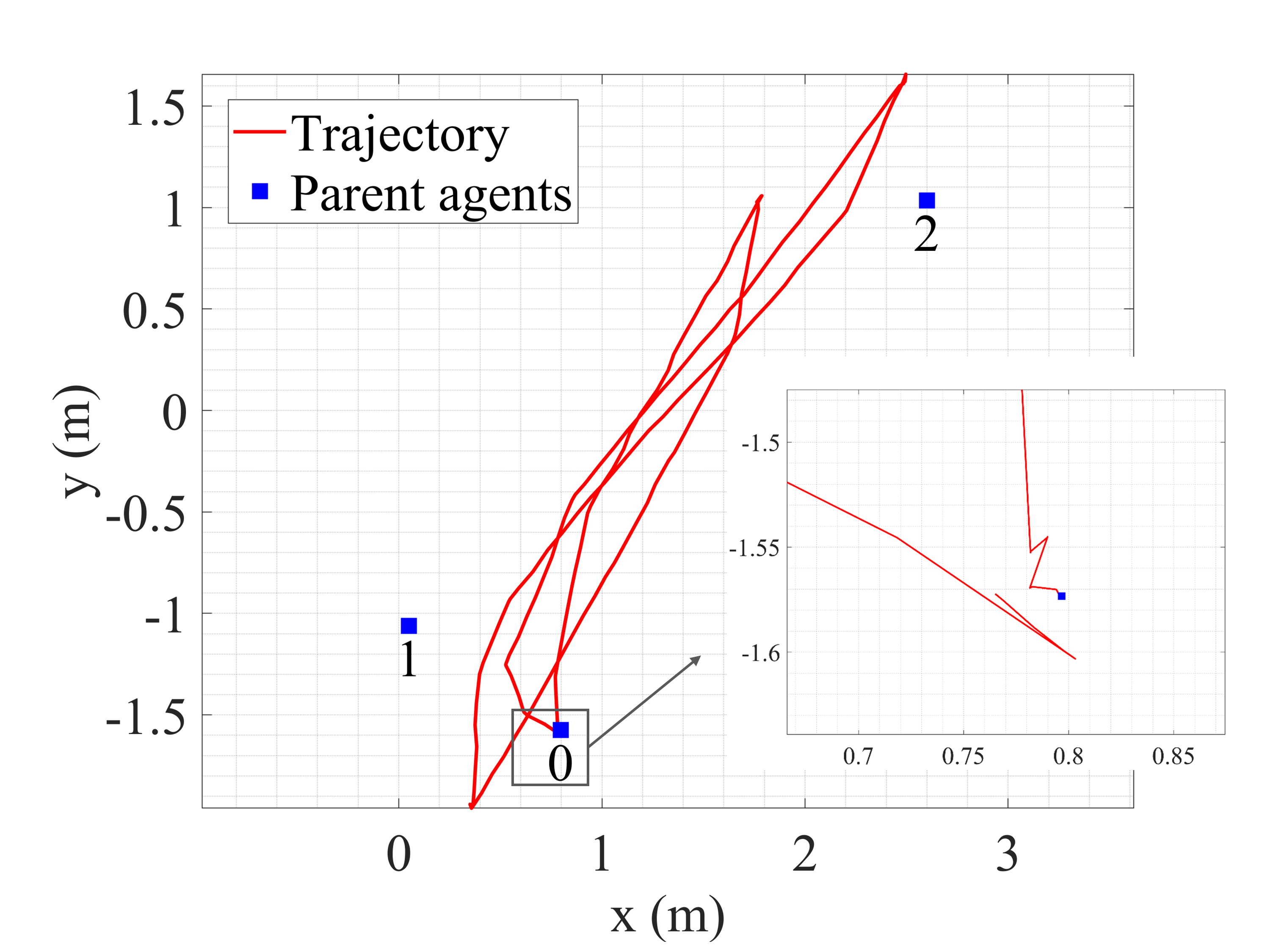}
    \label{fig:exp1_setup2}} 
    \caption{(a) Three parent agents equipped with our UWB hardwares.
    (b) The topology of the parent agents and the trajectory of the UGV obtained using a LiDAR sensor.} 
    \label{fig:exp1_setup}
\end{figure}

\textit{1) Setup:}
We equip three parent agents with our UWB hardwares, as shown in 
\mbox{Fig. \ref{fig:exp1_setup1}}. 
One of the parent agents is an unmanned ground vehicle (UGV) that is manually controlled to randomly 
move in a rectangular area of \mbox{2.5 m $\times$ 4 m}, with a maximum velocity 
of \mbox{0.25 m/s}. The rest of them are set as static parent agents. 
For performance comparison, the UGV is equipped with a LiDAR and a LiDAR map of the environment is 
established in advance.
The UGV's position is then tracked by finding the correspondences to the map with a frequency of 2 Hz.
The standard deviation of the 2D ground truth position error is approximately 0.05 m. 
Since the static agents are represented as 
obstacles in the LiDAR map, their positions are then directly obtained by querying the map.
\mbox{Fig. \ref{fig:exp1_setup2}} illustrates the topology of the parent agents and the output trajectory of the UGV.
In this way, by logging the output positions, we can obtain the real-time 
distance between parent agents, which is collected as ground truth. 
Further, the UGV is time synchronized with the static agents 
by our hardware setup sending one-pulse-per-second (1PPS)
timing packets.

We then utilize a virtual parent agent, which only receives the broadcast UWB packets but 
runs a distributed inter-agent range estimator at 100 Hz. The distances between the three 
parent agents are real-time logged to a PC for evaluation.
They are linearly fitted to the ground truth to calibrate the bias caused by antenna delays in 
(\ref{eq:twrwithbias}).

\begin{figure*}[h]
    \centering
    \subfloat[]{\includegraphics[width=0.32\linewidth]{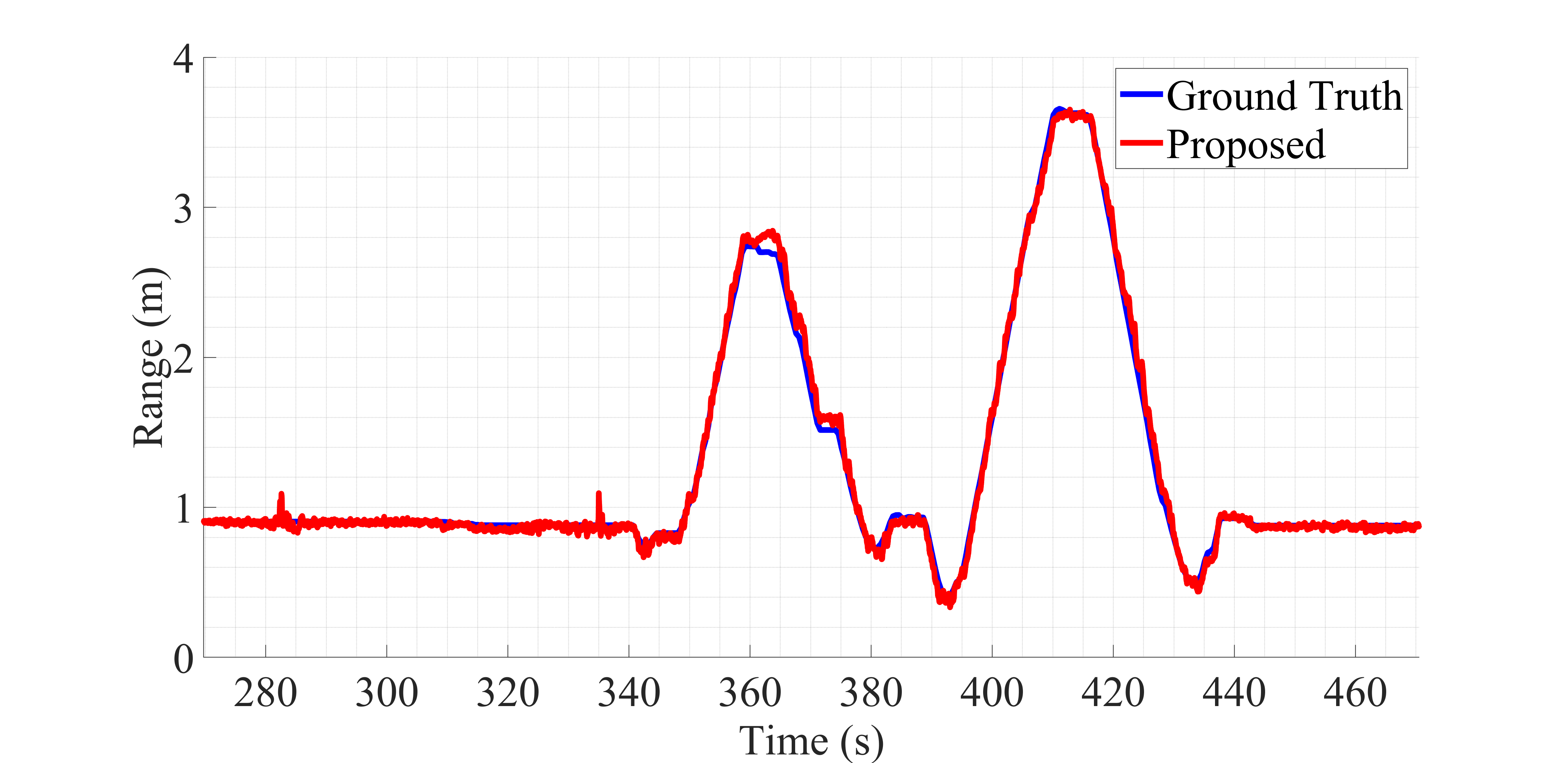}
    \label{fig_first_case}}
    \hfil
    \subfloat[]{\includegraphics[width=0.32\linewidth]{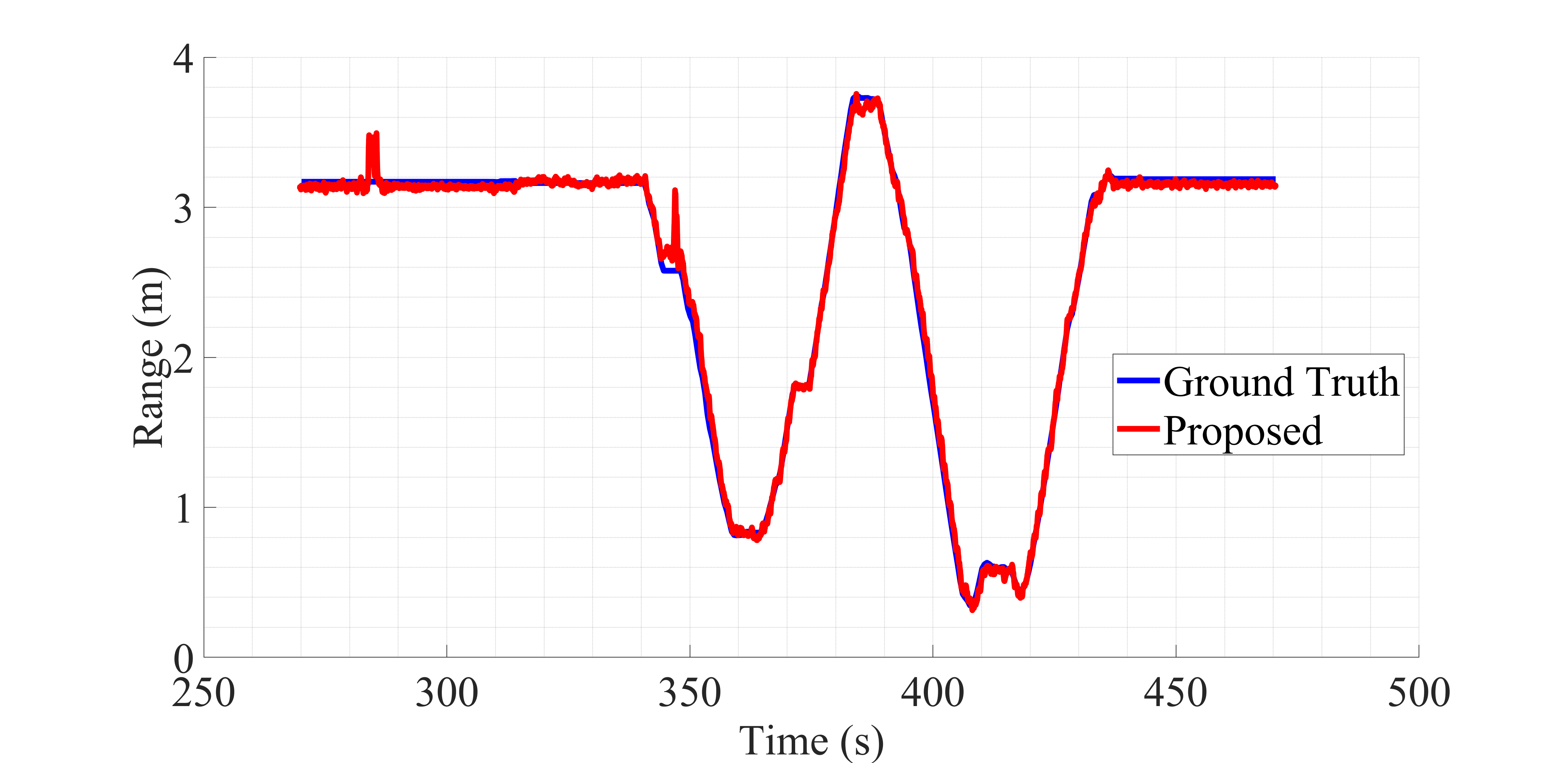}
    \label{fig_second_case}}
    \hfil
    \subfloat[]{\includegraphics[width=0.32\linewidth]{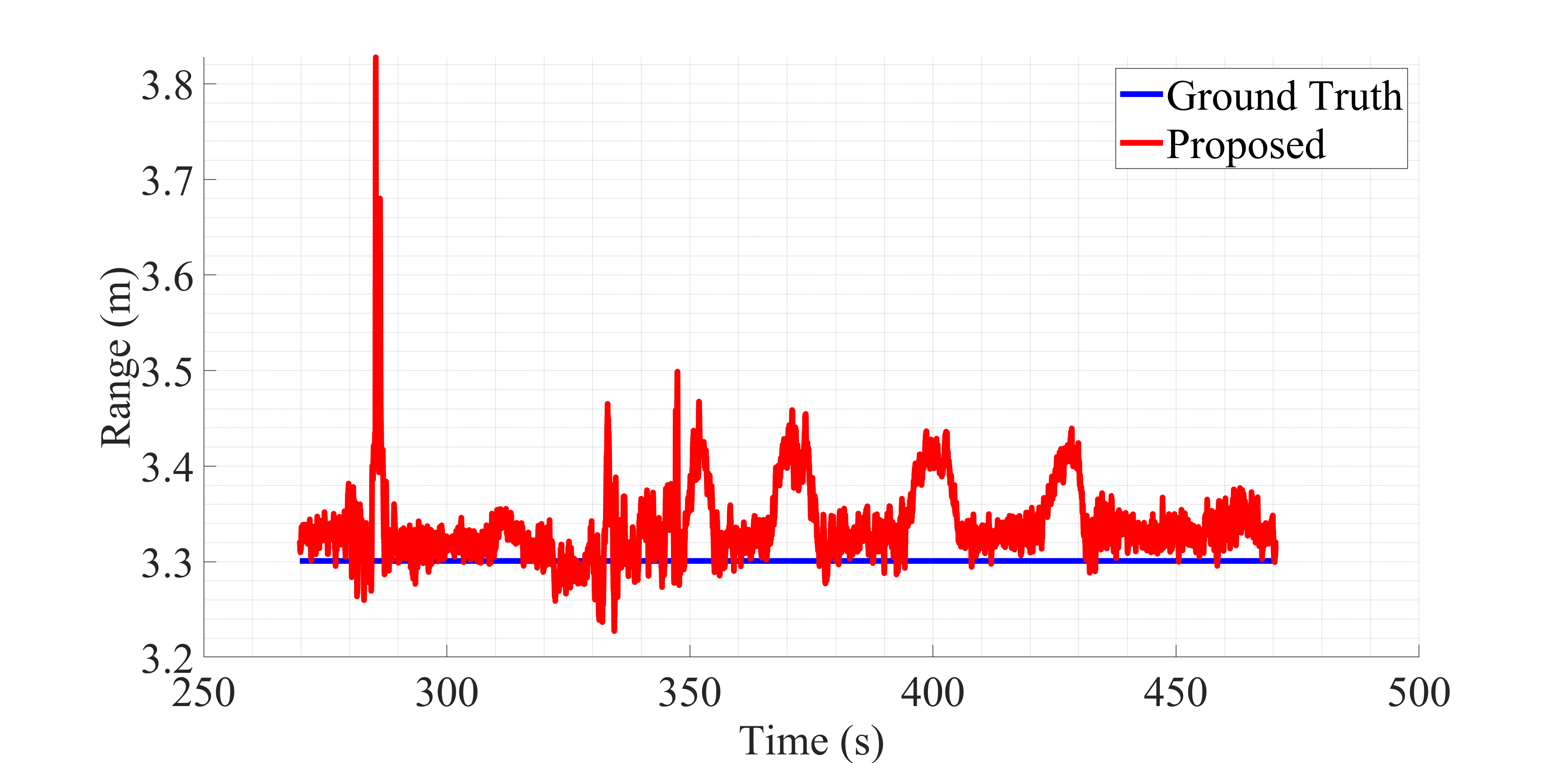}
    \label{fig:derror12}}
    \caption{The inter-agent ranging results for each pair of parent 
    agents. The red solid line indicates the estimation results while the 
    blue solid line indicates the ground truth. (a) The range between parent 
    agent $0$ and $1$, $d^{01}$.
    (b) The range between parent 
    agent $0$ and $2$, $d^{02}$. (c) The range between parent 
    agent $1$ and $2$, $d^{12}$.}
    \label{fig:twr results}
\end{figure*}

\begin{figure*}[h]
    \centering
    \subfloat[]{\includegraphics[width=0.32\linewidth]{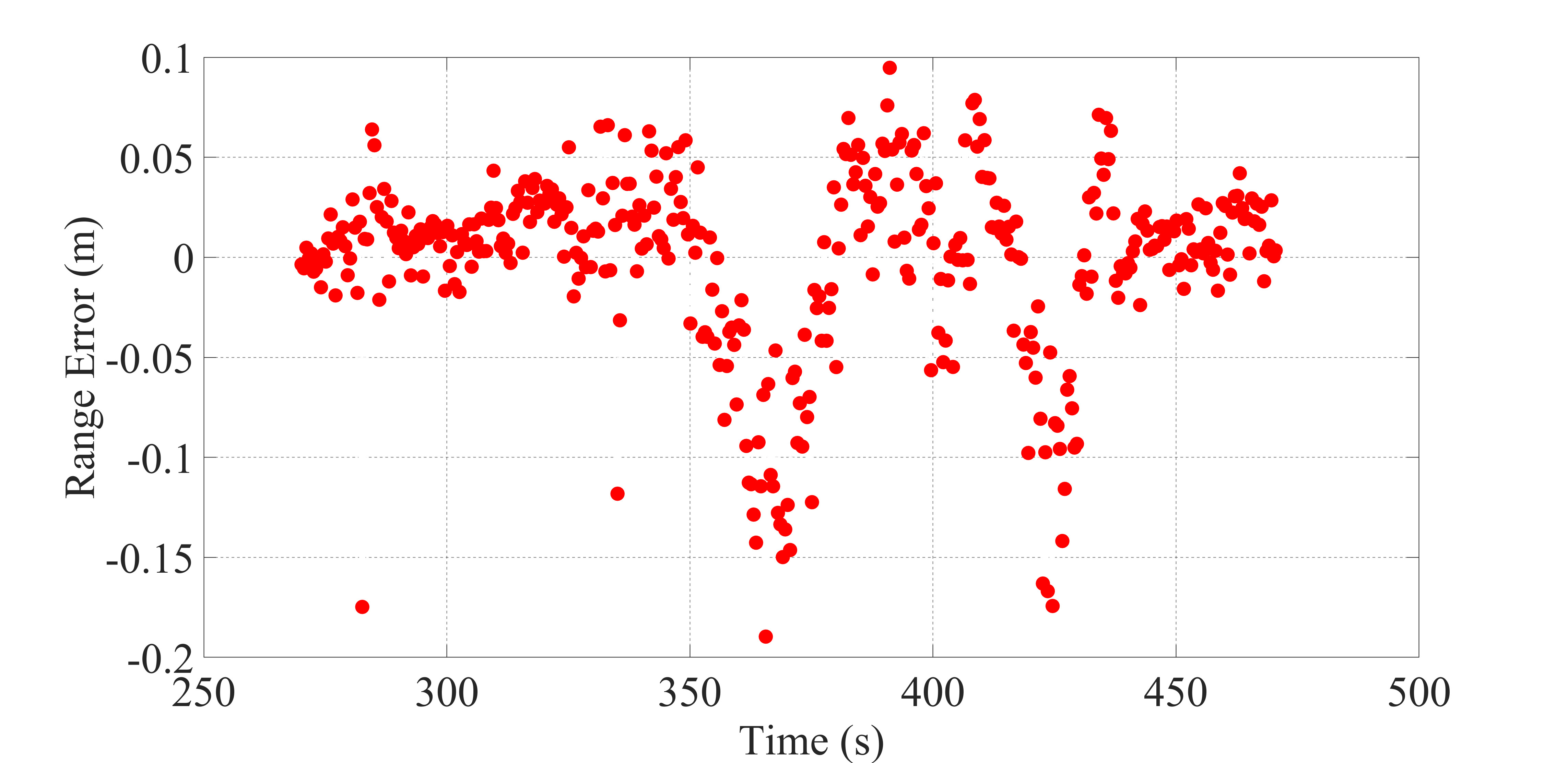}
    \label{fig_first_case1}}
    \hfil
    \subfloat[]{\includegraphics[width=0.32\linewidth]{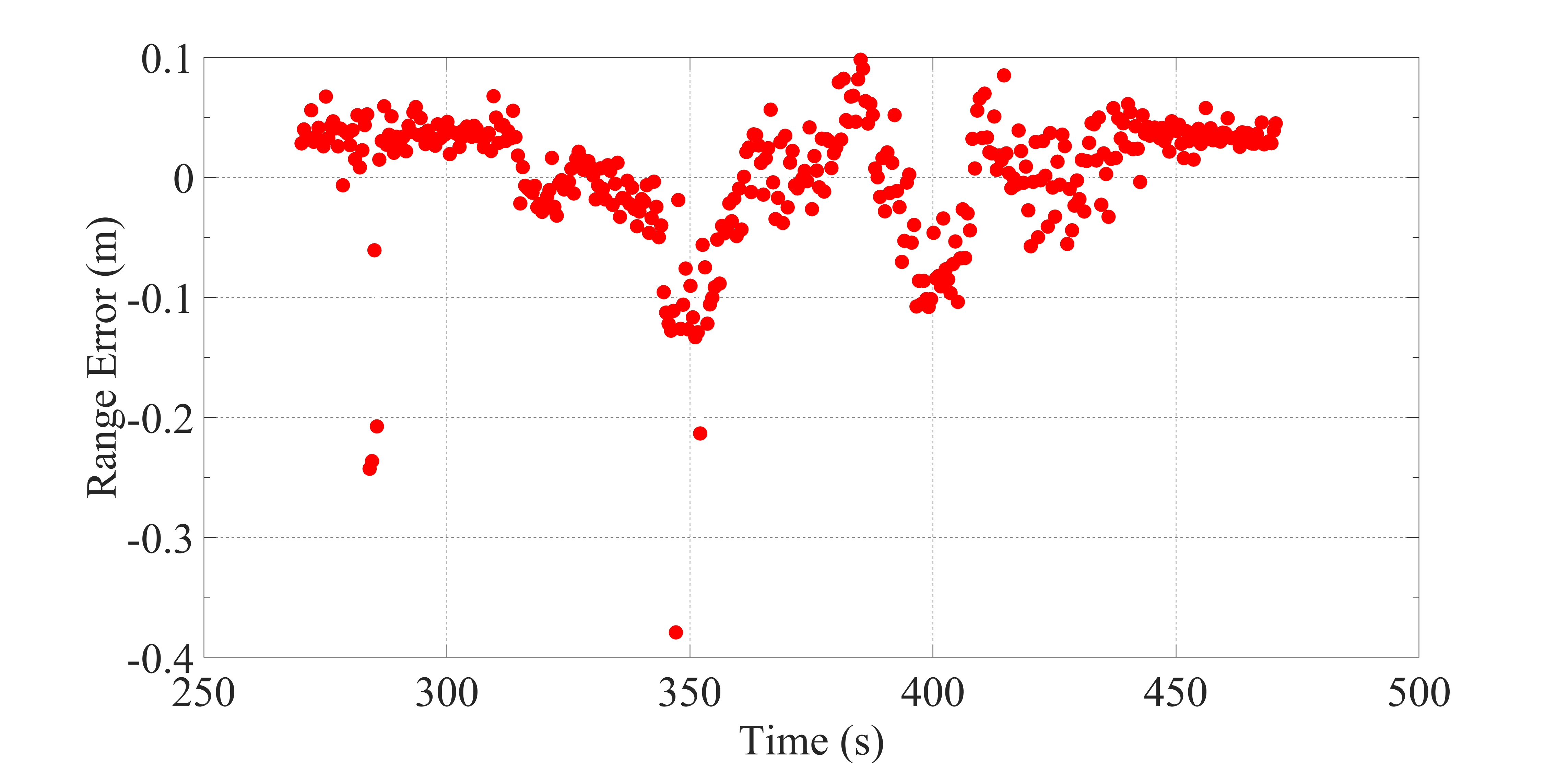}
    \label{fig_second_case2}} 
    \hfil
    \subfloat[]{\includegraphics[width=0.32\linewidth]{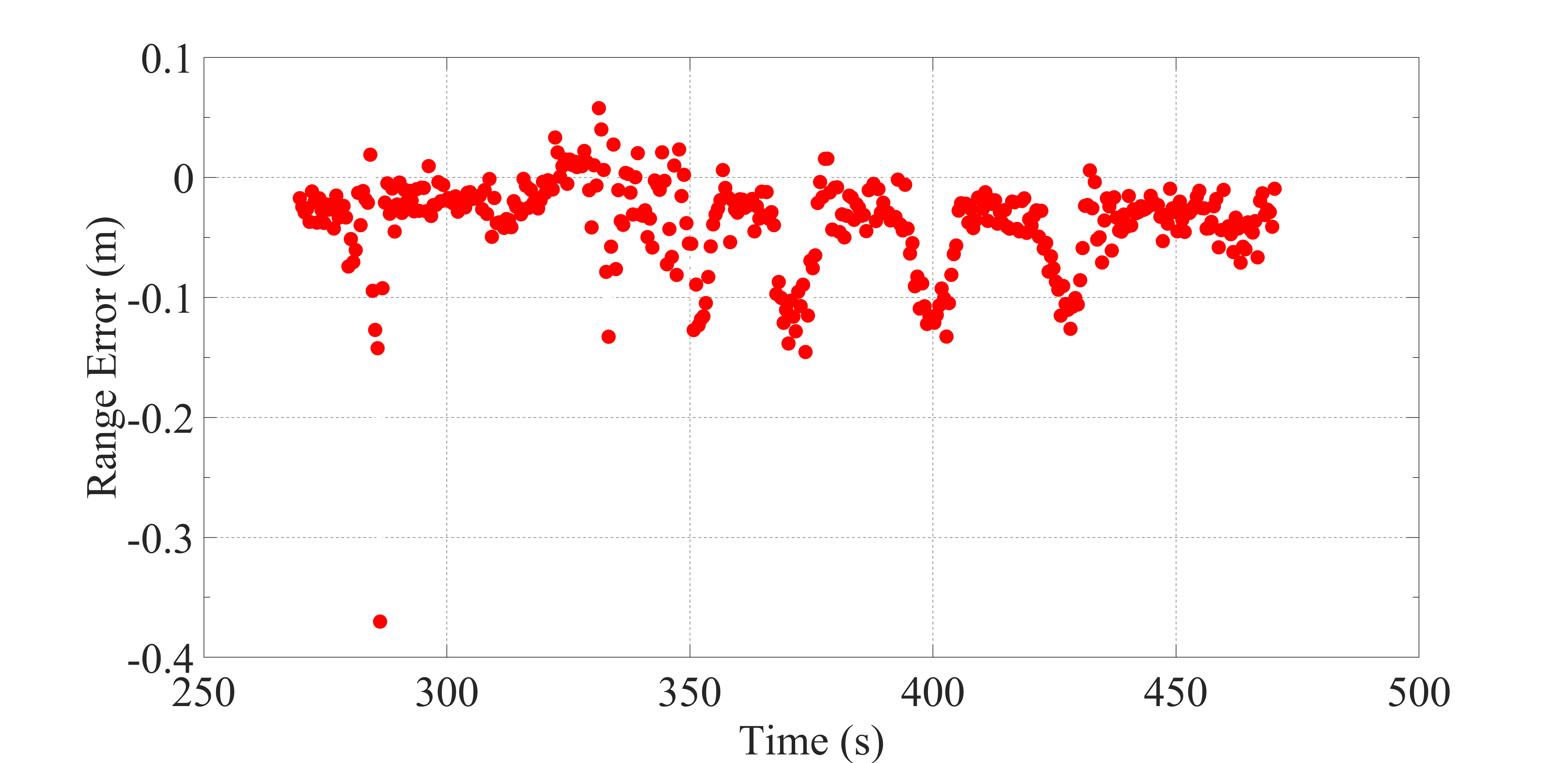}
    \label{fig_second_case3}}
    \caption{The error of inter-agent range for each pair of parent agents. 
    (a) Parent agent $0$ and $1$.
    (b) Parent agent $0$ and $2$. 
    (c) Parent agent $1$ and $2$.
    The average ranging error is 0.051 m.}
    \label{fig:twr error}
\end{figure*}

\textit{2) Results:}
\mbox{Fig. \ref{fig:twr results}} illustrates the inter-agent ranging results for 
three parent agents. As we can see, the estimated range is quite close to the 
ground truth. Note that the repetitive large range noise in \mbox{Fig. \ref{fig:derror12}}
is caused by the NLOS effects, in which case the UGV moves 
through the line connecting two static parent agents (see \mbox{Fig. \ref{fig:exp1_setup2}}). 
The ranging error is demonstrated in \mbox{Fig. \ref{fig:twr error}}. 
In terms of RMSE, the ranging results of our method yields an average accuracy of 0.051 m 
and the associated standard deviation is \mbox{0.050 m}. 
The ranging accuracy corresponds to a clock synchronization accuracy of 
about \mbox{0.172 ns}, which satisfies the synchronization requirements for most MASs.


\subsection{Child agent experiments}
Previous experiments have shown good quality of the ranging results. Consequently,
good clock synchronization and relative localization results are obtained, 
and a spatiotemporal reference is then established. To demonstrate that the JLAS of 
child agents is also applicable to dense, dynamic and real-time child agents, we utilize 
four static parent agents to establish a static spatiotemporal reference. For any number 
of child agents, we then try to synchronize to and localize against the spatiotemporal 
reference. Since the clock parameters have no ground truth, we focus to evaluate the 
localization performance. 
\begin{figure}
    \centering
    \includegraphics[width=0.9\linewidth]{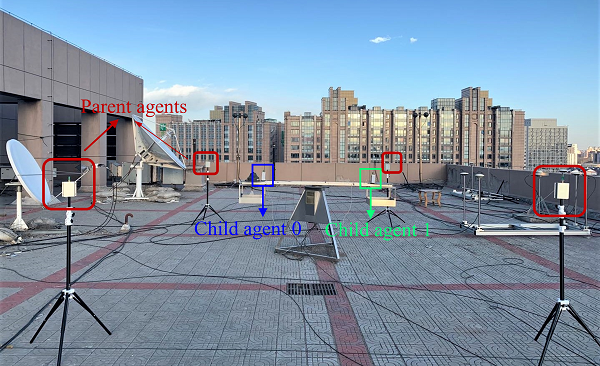} 
    \caption{The topology of the agents in our experiments. Four static parent agents are set on ground (red) and 
    two child agents are mounted on the two ends of a turntable (blue and green).}
    \label{fig:exp2}
\end{figure}

\textit{1) Setup:}
Four static parent agents are set on flat ground with the same heights. They form 
an approximate \mbox{5 m $\times$ 7 m} rectangular, and their relative positions are post-processed 
using their inter-agent ranging results. Two child agents are mounted on the two ends of a
turntable within the rectangular formed by parent agents, as illustrated in \mbox{Fig. \ref{fig:exp2}}.
The turntable has a diameter of 1.733 m, and rotate 
at an average angular velocity of \mbox{0.14 rad/s}. 
The two child agents then move with a linear velocity of \mbox{0.12 m/s} 
and localize themselves by only receiving the UWB packets.
The localization results are real-time logged to a host computer for 
evaluation. Since we have no ground truth of child agent positions, we use the error of 
distance between two child agents as main evaluation metric. 
The distance is calculated using the estimated localization results.

\begin{figure*}[!t]
    \centering
    \subfloat[]{\includegraphics[width=0.32\linewidth]{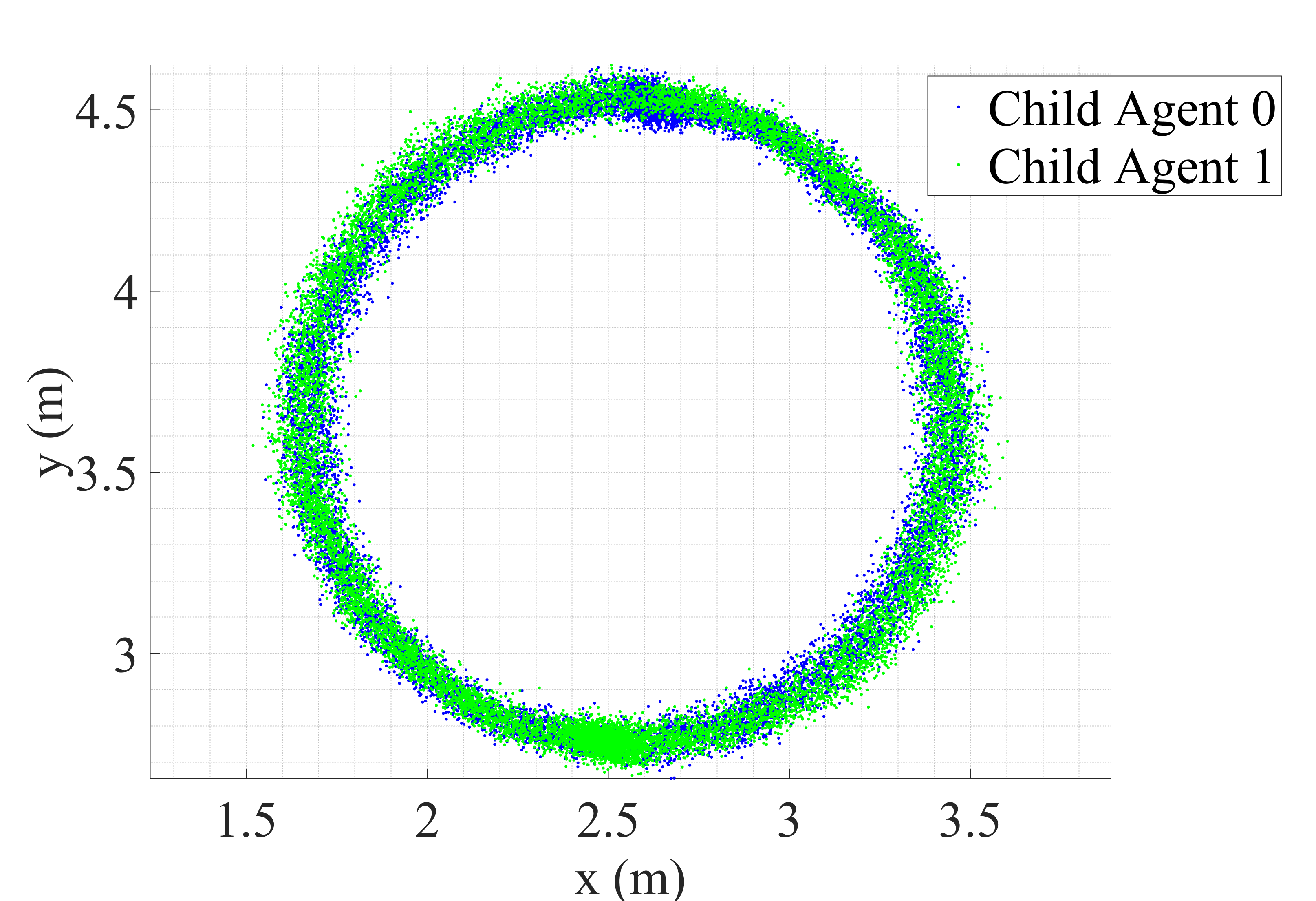}
    \label{fig:circle}}
    \hfil
    \subfloat[]{\includegraphics[width=0.32\linewidth]{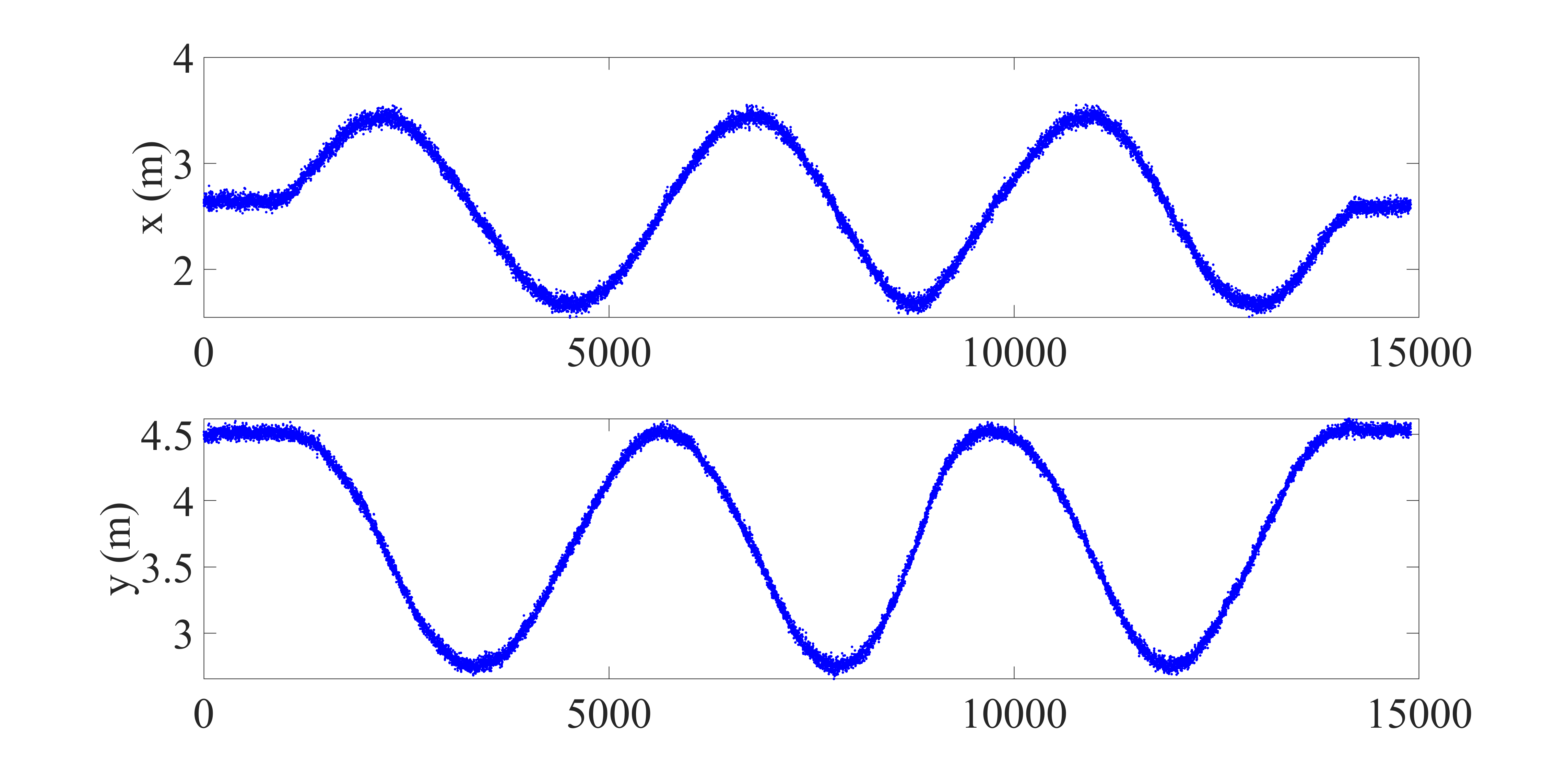}
    \label{fig:ca1}} 
    \hfil
    \subfloat[]{\includegraphics[width=0.32\linewidth]{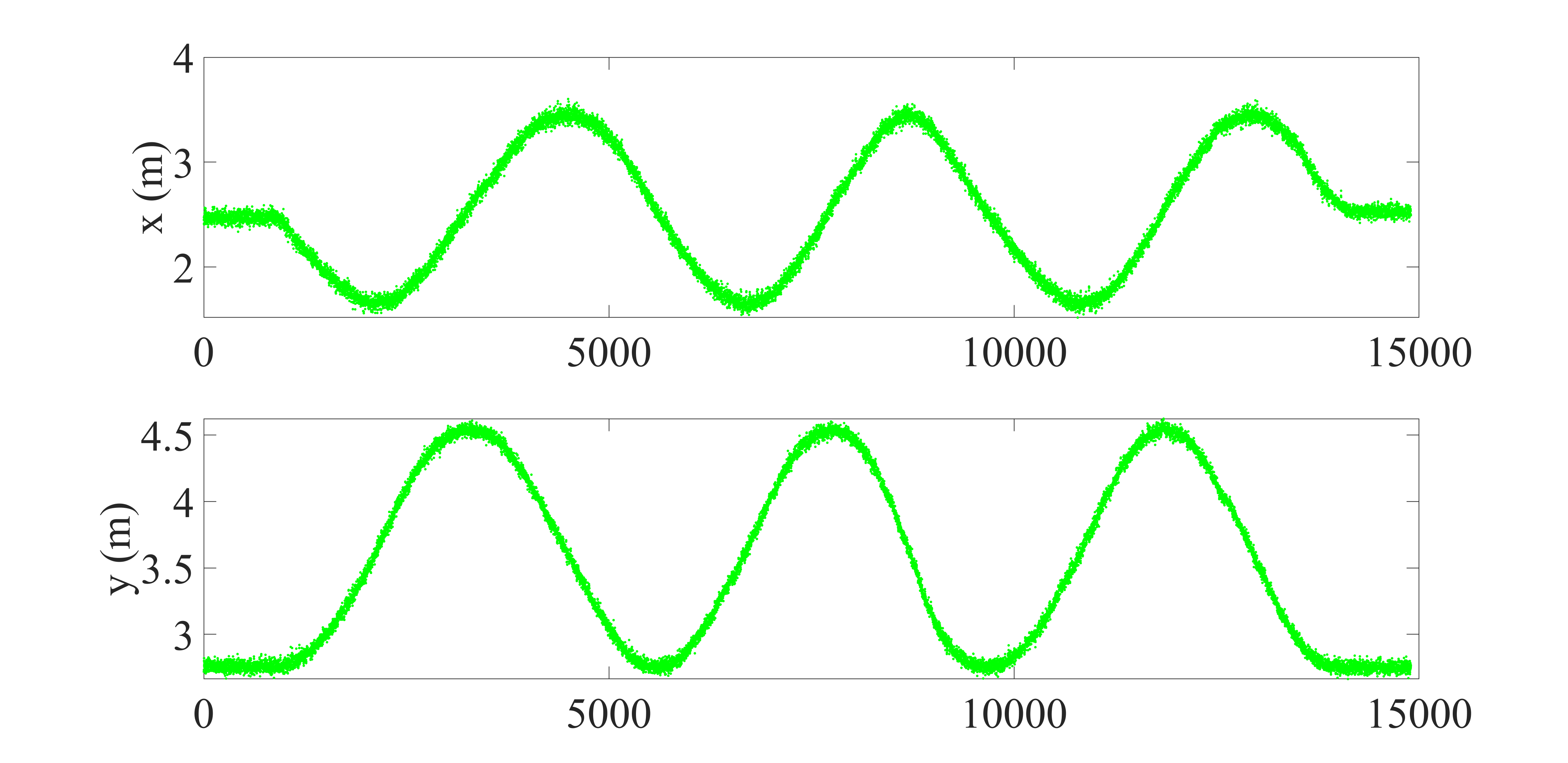}
    \label{fig:ca2}}
    \caption{Localization results for child agents. 
    (a) Estimated trajectories of child agent 0 (blue) and child agent 1 (green).
    (b) Estimated x y positions of child agent $0$. 
    (c) Estimated x y positions of child agent $1$.}
    \label{fig:child agent localization results}
\end{figure*}
\begin{figure}
    \centering
    \includegraphics[width=0.9\linewidth]{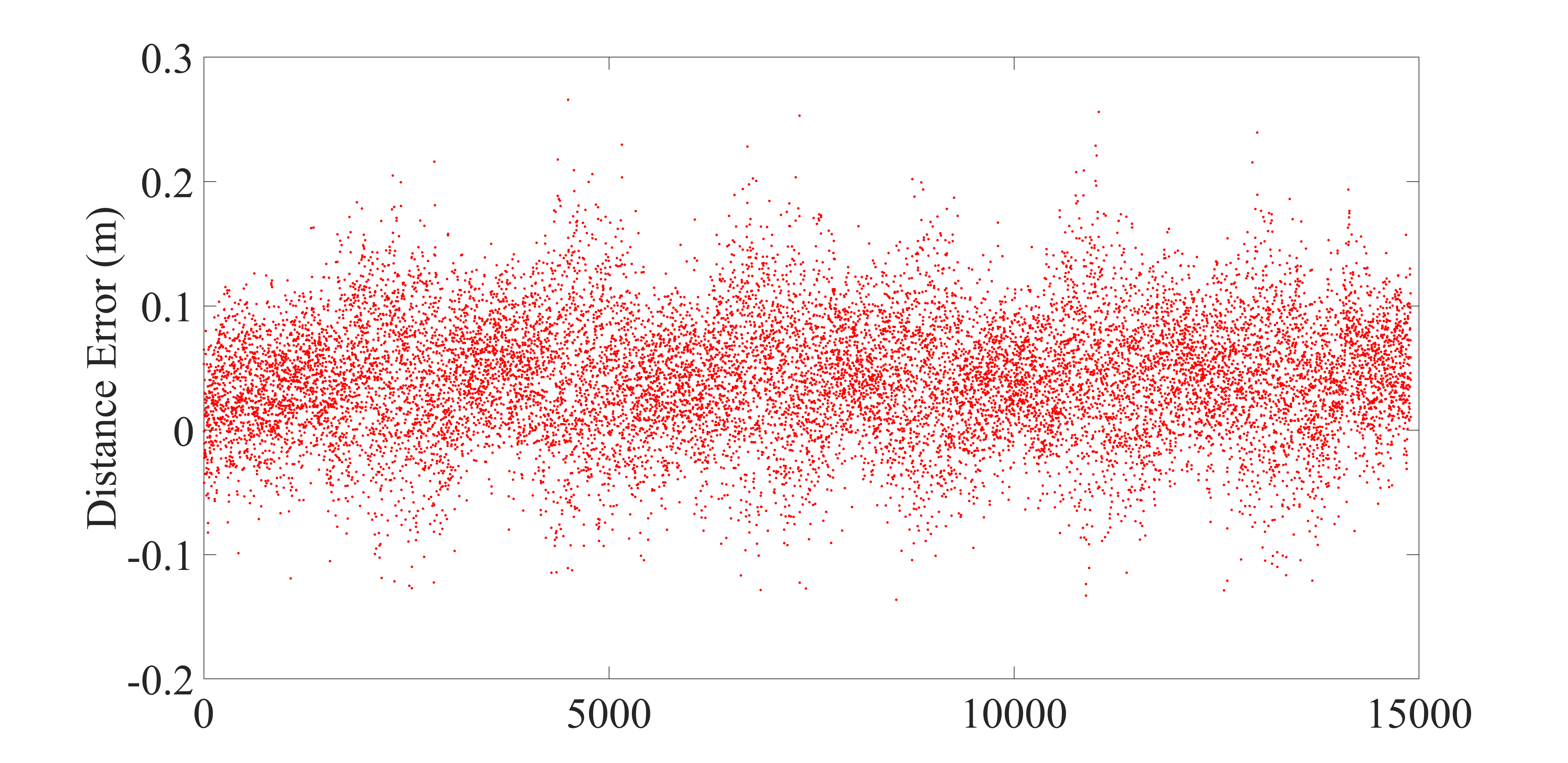} 
    \caption{The error of the calculated distance between child agent $0$ and $1$. The 
    distances are calculated using the estimated localization results.
    The average inter-agent range error is 0.064 m.}
    \label{fig:child agents error}
\end{figure}

\textit{2) Results:}
\mbox{Fig. \ref{fig:child agent localization results}} shows the relative localization results 
for two child agents. 
In \mbox{Fig. \ref{fig:circle}} the blue and green dotted lines indicate the estimated positions which are 
circular. 
Since the child agents are rigidly mounted on the two ends of the turntable, 
the coincident of the trajectories provides strong qualitative evidence of the 
algorithm's performance in accuracy and consistency. 
\mbox{Fig. \ref{fig:ca1}} and \ref{fig:ca2}
illustrate the estimated x y positions, which are sinusoidal forms and 
present exact correspondence of the circular trajectories.
The distance between the two child agents are then calculated using the estimated
localization results, and the distance error is illustrated in \mbox{Fig. \ref{fig:child agents error}}.
The RMSE and the associated standard deviation of the calculated distance 
is \mbox{$0.064$ m} and \mbox{$0.048$ m}. The distance error appears to be dependent on the 
positions of the child agents. As we observe a relatively strong correlation 
between the repeatability of a bias on the order of 0.1 m contained in distance error 
and the circular pattern of movements for child agents. 
The bias is a determined value, thus can be calibrated in future works.
The main cause of the bias can be explained by the relative position and orientation between 
UWB antennas. 

\subsection{Discussion}
The experimental results for parent and child agents together show that we can 
obtain accurate and real-time relative localization results for a MAS. 
A typical nanosecond accuracy for clock synchronization can be achieved.
For high-density MASs containing a number of agents, we can choose a small set of them as 
parent agents performing packet broadcasting, as described in our broadcast architecture.
Our APBR protocol ensures the collision-free 
broadcasting, thus enabling efficient information sharing in parent agents to establish 
a real-time updated spatiotemporal reference, as shown in the parent agent experimental results. 
The rest of the agents are then set as child agents that only receive the 
broadcast packets. In this way, the system can simultaneously support theoretically an unlimited 
number of child agents. The child agent experiments have shown that we can real-time and accurately
localize the child agents against the spatiotemporal reference using the received packets.
Though we only show the results for two child agents, the 
extension to any number of child agents is straightforward if the NLOS effects can be neglected.
In our experiments, the broadcasting frequency is set as 100 Hz. As our protocol supports 
JLAS of child and parent agents using the same broadcast packets and our 
estimation approaches are distributely implemented, the estimation frequency for both 
child and parent agents
is then 100 Hz. Therefore, our system can also be applied to high-maneuverability MASs which require 
real-time and high-frequency spatiotemporal state estimation.
Further, our system is developed into a stand-alone measurement unit with low weight, small volume 
and low power consumption, thus can be integrated into SWaP constrained MASs. 
To this end, our proposed BLAS system is proven feasible to be applied to high-density, high-maneuverability 
and SWaP constrained MASs for accurate, high-frequency and real-time relative localization and clock synchronization.

\textbf{Remark} Comparisons between our BLAS with the state-of-arts are not straightforward,
since the previous relevant methods are not specially designed for 
a dynamic and dense MAS in an infrastructure-free environment as we have discussed in Section II.
To the best of our knowledge, we are the first to address the JLAS problem for such 
a MAS using a broadcast UWB architecture. 
 
\section{Conclusion}
In this paper, we present a wireless broadcast relative localization and clock synchronization system
for MASs with high-density, high-maneuverability and SWaP constraint characteristics.
A broadcast architecture and the supporting approaches, such as an ABPR communication protocol and distributed 
parent-child state estimators have been presented and implemented in our system.
Simulation and experimental results verified that the proposed BLAS system is capable of establishing accurate and high-frequency
relative localization and clock synchronization for dynamic dense MASs with limited resources.

The proposed system can be applied to a number of scenarios, e.g., establishing a UAV Ad-Hoc network, 
in which parent agents can
output real-time topology and clock parameters of the dynamic network to aid network routing.
Additionally, although this study focuses on the localization results which are relative, it 
can be extended to obtain absolute localization for MASs. In that case, our BLAS system can be used 
as a sensory system, which yields ranging measurements between parent agents.
By utilizing other 
proprioceptive sensors, such as cameras, LiDARs or IMUs, we can utilize 
collaborative localization algorithms to establish absolute spatiotemporal references in 
parent agents. The parent agents can then serve like pseudo-GPS satellites, which enables any number of 
users (child agents) to retrieve their spatiotemporal information in unknown environments.

As part of future work, we will extend the proposed BLAS to 3D space and include more extensive 
evaluation of the proposed methods.


%


\ifCLASSOPTIONcaptionsoff
    \newpage
\fi



\bibliographystyle{IEEEtran}
\bibliography{IEEEabrv,paper}
%


%








\end{document}